\documentclass[10pt,twocolumn,letterpaper]{article}

\usepackage[pagenumbers]{cvpr} 

\definecolor{cvprblue}{rgb}{0.21,0.49,0.74}
\usepackage[pagebackref,breaklinks,colorlinks,allcolors=cvprblue]{hyperref}

\usepackage{makecell}
\usepackage{graphicx}
\usepackage{booktabs}
\usepackage{multirow}
\usepackage{tcolorbox}
\usepackage{pifont}
\usepackage{multicol}

\definecolor{Gray}{gray}{0.9}
\definecolor{demphcolor}{RGB}{144,144,144}
\definecolor{backblue}{RGB}{221,239,251}
\definecolor{backblue1}{RGB}{186,216,242}
\definecolor{backred}{RGB}{244,199,204}

\definecolor{original}{RGB}{153, 153, 255}
\definecolor{amodif}{RGB}{255, 153, 153}

\definecolor{darkblue}{RGB}{37, 52, 148}
\definecolor{lightgreen}{RGB}{216, 240, 211}

\newcommand{\best}[1]{\vspace{-2.4mm}\textbf{\colorbox{backblue1}{\makebox(18,5){\vspace{0.15mm}#1}}}}
\newcommand{\second}[1]{\vspace{-2.4mm}\colorbox{backblue}{\makebox(18,5){\vspace{0.15mm}#1}}}
\newcommand{\std}[1]{\scriptsize{$\pm$#1}}

\definecolor{softgray}{rgb}{0.9, 0.9, 0.9}

\definecolor{my_green}{RGB}{51,102,0}
\newcommand{\cmark}{\textcolor{my_green}{\ding{51}}} 

\newcommand{\method}{\textsc{NEMO}}
\newcommand{\methodemojititle}{\textsc{\method}\includegraphics[width=0.034\textwidth]{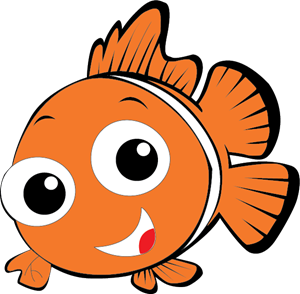}\xspace}

\usepackage{pifont}
\usepackage{colortbl}
\usepackage{xcolor}
\usepackage{array} 
\usepackage{nicematrix}
\usepackage{mdframed}
\usepackage{float}
\usepackage{longtable}

\def\confName{CVPR}
\def\confYear{2025}

\title{\methodemojititle: Can Multimodal LLMs Identify Attribute-Modified Objects?}

\author{Jiaxuan Li$^{1}$,\ \ Junwen Mo$^{1}$,\ \ Duc Minh Vo$^{1}$,\ \ Akihiro Sugimoto$^{2}$,\ \ Hideki Nakayama$^{1}$\\
$^{1}$The University of Tokyo, Japan\ \ $^{2}$National Institute of Informatics, Japan \\
\tt\small \{li, mo, vmduc\}@nlab.ci.i.u-tokyo.ac.jp\ \ sugimoto@nii.ac.jp\ \ nakayama@ci.i.u-tokyo.ac.jp\\
}

\begin{document}

\addtolength{\baselineskip}{-0.38pt}

\twocolumn[{%
\renewcommand\twocolumn[1][]{#1}%
\maketitle
\begin{center}
    \centering
    \captionsetup{type=figure}
    \includegraphics[width=0.99\textwidth]{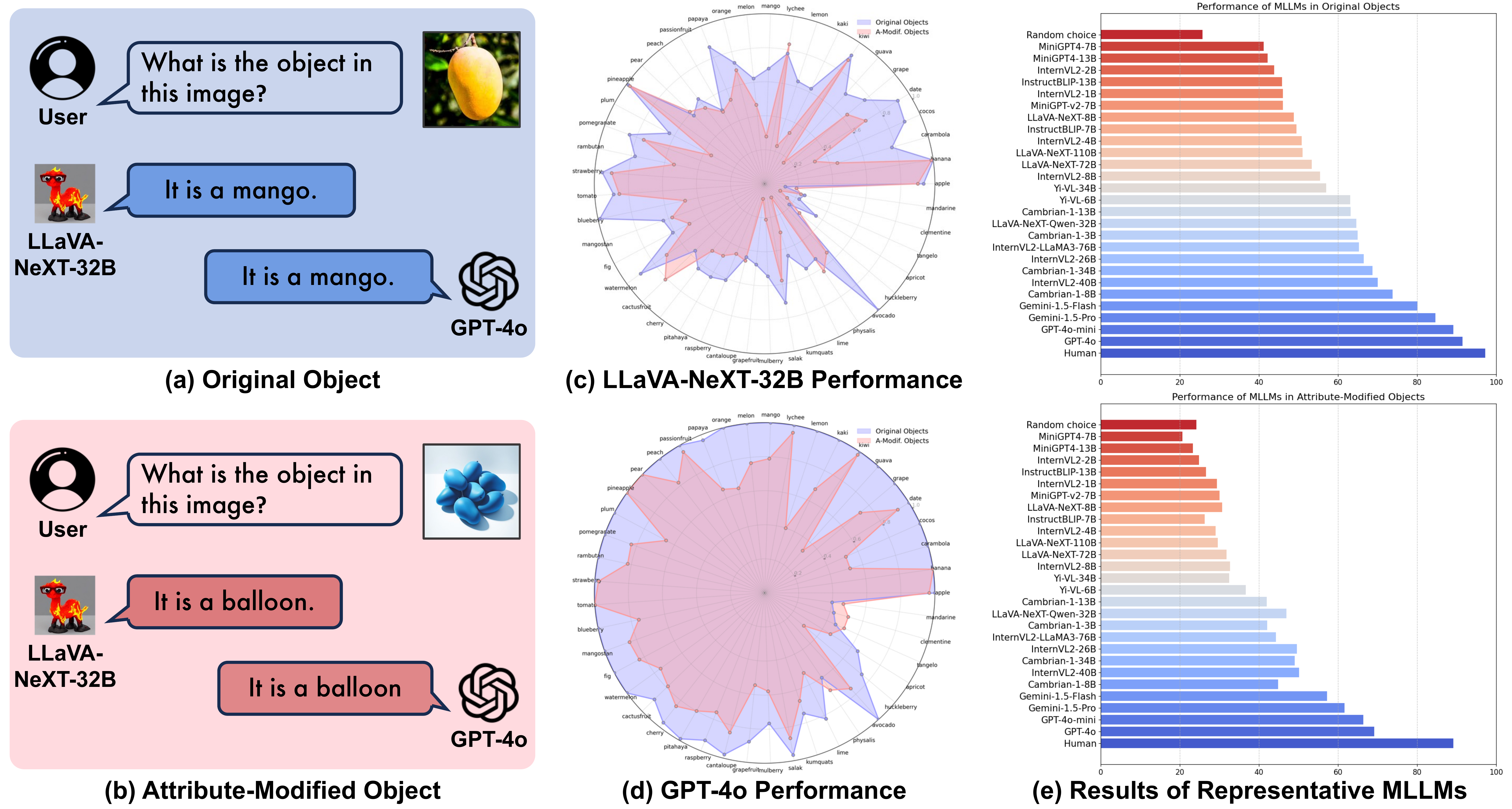}
    \captionof{figure}{Challenges in consistently identifying original and attribute-modified objects are illustrated in (a) and (b): while LLaVA-NeXT-Qwen-32B and GPT-4o correctly recognize the original \textit{``mango"}, both fail with a blue variant. Panels (c) and (d) show average accuracy scores of LLaVA-NeXT-Qwen-32B and GPT-4o across objects in our \method{}, comparing \textcolor{original}{original} and \textcolor{amodif}{attribute-modified} versions. (e) provides a comparative overview of average scores for representative MLLMs on original (upper) and attribute-modified (lower) objects. 
    }
    \label{fig1_teaser}
\end{center}%
}]

\begin{abstract}

Multimodal Large Language Models (MLLMs) have made notable advances in visual understanding, yet their abilities to recognize objects modified by specific attributes remain an open question.
To address this, we explore MLLMs' reasoning capabilities in object recognition, ranging from commonsense to beyond-commonsense scenarios.
We introduce a novel benchmark, \method{}\footnote{Our \method{} and evaluation code are available at \url{https://jiaxuan-li.github.io/NEMO}.},  which comprises 900 images of origi\underline{N}al fruits and their corresponding attribut\underline{E}-\underline{MO}dified ones; along with a set of 2,700 questions including open-, multiple-choice-, unsolvable types.
We assess 26 recent open-sourced and commercial models using our benchmark.
The findings highlight pronounced performance gaps in recognizing objects in \method{} and reveal distinct answer preferences across different models.
Although stronger vision encoders improve performance, MLLMs still lag behind standalone vision encoders.
Interestingly, scaling up the model size does not consistently yield better outcomes, as deeper analysis reveals that larger LLMs can weaken vision encoders during fine-tuning.
These insights shed light on critical limitations in current MLLMs and suggest potential pathways toward developing more versatile and resilient multimodal models.

\end{abstract}  

\section{Introduction}
\label{sec:intro}

\noindent \textit{Our world is constantly changing, and with it comes the possibility of unexpected events - such as all mango skins suddenly turning blue (Fig.~\ref{fig1_teaser}).
With prior knowledge, humans would still recognize these ``blue mango" despite the modified color.
Yet, \textbf{could MLLMs do the same when objects are modified by beyond-commonsense attributes}?} --- This question is the motivation of this work.

Multimodal Large Language Models (MLLMs)~\cite{gpt4v, gemini, liu2023llava, li2024llavanext-strong, chen2024far, tong2024cambrian, internlmxcomposer2_5, ye2023mplugowl} have demonstrated remarkable capabilities in various visual understanding tasks, benefiting from the extensive world knowledge embedded within Large Language Models (LLMs)~\cite{touvron2023llama,qwen2023, cai2024internlm2}.
As MLLMs continue to evolve, a thorough exploration of their visual understanding and reasoning capabilities is essential, as it paves the way for developing more reliable and robust multimodal models.
A wide range of benchmarks have been developed, focusing on basic perception tasks~\cite{liu2023llava, fu2023mme, li2024seed, li2023seedbench2}, as well as more specialized challenges within commonsense, such as reasoning under adversarial conditions~\cite{naturalbench, zhang2024avibench, zhao2023adv} and mitigating hallucination~\cite{wang2024haloquest, li2023pope, guan2024hallusionbench}.
While prior studies have extensively addressed commonsense scenarios~\cite{bittonguetta2024visualriddles, li2024foodieqa, li2024seed, li2023seedbench2, yue2023mmmu, yue2024mmmupro}, our work fully ventures into the area of MLLM behavior as it shifts from commonsense to beyond-commonsense reasoning.
This is because, as we stated above, our world always changes and the beyond-commonsense \textit{``blue mango"} resulting from mutation may come out.
Developing MLLMs to handle such novel situations is crucial in advancing them to human-level AI~\cite{mccarthy2007here, morris2024agi}.
As a first step, this study aims to comprehensively assess existing MLLMs to answer our motivation question while giving insights for developing more versatile and resilient models.

We begin by evaluating MLLMs' performance concerning the transition from recognizing \textbf{original objects} to \textbf{attribute-modified objects}, to understand how well they adapt when object attributes deviate from commonsense expectations.
Focusing on beyond-commonsense reasoning is not entirely new: as summarized in Tab.~\ref{tab:benchmarks}, overarching objectives of a few benchmarks~\cite{tai2024link,Bitton-Guetta_2023_WHOOPS,zhou2023rome} are closely aligned with ours.
However, attribute-modified object identification remains largely under-explored. 
These benchmarks do not provide sufficient insight into MLLMs' capabilities when facing our scenario.
Furthermore, they cannot be directly employed in our work due to the following reasons.
(i) They lack a set of standard original objects for comparision.
Standard original objects enable the model's responses to be assessed based on how well it identifies the object when its attributes change.
Without these reference objects, it becomes challenging to determine if the model consistently understands both the original object and its variant.
(ii) Not all images in these benchmarks contain a single object with a clearly modified attribute, yet it is still identifiable for humans based on their prior knowledge and adaptable perception ability. This point is essential to evaluate the model effectively.
(iii) None of them provide well-constructed VQA annotations to comprehensively test if the model can identify objects.

To address above limitations, we introduce \method{}, a benchmark designed to systematically evaluate MLLMs' reasoning from commonsense to beyond-commonsense scenarios.
\method{} includes 900 images and 2,700 carefully designed VQA annotations, covering 45 original objects and their attribute-modified counterparts.
We choose fruits as our target objects due to their easily modifiable attributes and clear recognizability, addressing limitation (ii).
Original objects are manually collected from the internet, while attribute-modified objects are generated by altering colors using DALL-E3~\cite{Aditya2021dalle}, addressing limitation (i).
The VQA task in \method{} involves recognizing objects in open, multiple-choice, and unsolvable questions, testing MLLMs' ability to identify objects via their own knowledge, choose between correct and deceptive options, and reject all deceptive choices, thereby addressing limitation (iii).

As illustrated in Fig.~\ref{fig1_teaser}e, we evaluate a variety of open-sourced and commercial models, including GPT-4o~\cite{gpt4v} and Gemini-1.5-Pro~\cite{gemini}, and conduct a user study to measure human performance. Our main contributions and findings can be summarized as follows:

\begin{itemize}[leftmargin=*]
\item 
We provide \method{}, a new MLLM benchmark consisting of original objects and their attribute-modified counterparts, designed to fully assess the model from commonsense to beyond-commonsense.
Note that we also extend \method{} to cover shape-modified fruits and color-modified animals (\cref{subsec:other}), which is further detailed in the Supplemental Material (SM).

\item Extensive experiments on 26 MLLMs reveal significant performance gaps between original and attribute-modified object recognition, and between open-sourced and commercial MLLMs (\cref{fig1_teaser}c-e, more in \cref{subsec:overall}).

\item We analyze factors causing performance drop in MLLMs, providing insights for developing more robust MLLMs (\cref{subsec:model,subsec:data}):
(1) MLLMs with stronger vision encoders perform better overall, but performance falls short of that achieved by their vision encoders alone.
(2) Scaling model size does not consistently improve recognition of original or attribute-modified objects.
(3) Stronger vision encoders benefit the recognition performance of MLLMs, while larger LLMs will weaken the vision encoders during fine-tuning.
(4) MLLMs exhibit preferences in their responses for certain objects. 

\end{itemize}

\begin{table*}[tb]
\centering
\caption{Comparison of \method{} with recent visual understanding benchmarks for MLLMs, in terms of scale (VQA and image count), domain (real and synthetic), question type (yes/no, open, multiple-choice, and unsolvable questions), object type (fine-grained and attribute-modified), whether human or GPT is evaluated (H/G Eval.), evaluation objectives, and the number of evaluated models.}
\label{tab:benchmarks}
\setlength\tabcolsep{3pt}
\resizebox{\linewidth}{!}{
\begin{NiceTabular}{@{}l|cc|cc|cccc|cc|l|c|c@{}}
\toprule[1pt]
\multirow{2}{*}{\textbf{Benchmark}} & \multicolumn{2}{c}{\textbf{Scale}}  & \multicolumn{2}{c}{\textbf{Domain}} & \multicolumn{4}{c}{\textbf{Question Type}}  & \multicolumn{2}{c}{\textbf{Object Type}} & \multirow{2}{*}{\textbf{What to Evaluate}} & \multirow{2}{*}{\textbf{\#Models}} & \multirow{2}{*}{\textbf{H/G Eval.}} \\

\cmidrule(lr){2-3} \cmidrule(lr){4-5}  \cmidrule(lr){6-9} \cmidrule(lr){10-11}
&\#VQA &\#Img. & Real & Synth. & Y/N & Open & Choice & Unsol. &Orig. & Attr.-Mod. \\
\midrule

\multicolumn{12}{@{}l@{}}{\textbf{General Visual Understanding Evaluation}} \\
~~LLaVA-Bench~\cite{liu2023llava} &60 &24 & \cmark & & & \cmark & & &\cmark &  & Basic perception &4 &GPT  \\
~~MME~\cite{fu2023mme} & 1,457 & 1,187 & \cmark & & \cmark & & & &\cmark  &  & Basic perception &6 &GPT \\
~~MMBench~\cite{liu2025mmbench} &3,217 &3,217 & \cmark & & & & \cmark & &\cmark & &Basic perception &14 &GPT  \\
~~SEED-Bench-1~\cite{li2024seed} &19,242 &-- & \cmark & & & & \cmark & &\cmark & & Basic perception &18 &N/A\\
~~SEED-Bench-2~\cite{li2023seedbench2} &24,371 &-- & \cmark & & & & \cmark & &\cmark & & Basic perception &23 &N/A  \\
~~MMMU~\cite{yue2023mmmu} &11,500 &-- & \cmark & & & \cmark & \cmark & &\cmark &  &Expert-level domain knowledge &30 &Human\&GPT \\
~~MMMU-Pro~\cite{yue2023mmmu} &3,460 &-- & \cmark & & & \cmark & \cmark & &\cmark &  &Expert-level domain knowledge &21 &GPT \\
\midrule

\multicolumn{12}{@{}l@{}}{\textbf{Key Capability Evaluation within Commonsense}} \\
~~NaturalBench~\cite{naturalbench} &7,600 &-- & \cmark & & \cmark & & \cmark & &\cmark &  &Natural adversarial samples reasoning &53 &Human\&GPT \\
~~UPD~\cite{miyai2024unsolvable} &2,095 & -- & \cmark & & & & & \cmark &\cmark &  &Answer withholding  &7 &GPT \\
~~FoodieQA~\cite{li2024foodieqa} &403 &389 & \cmark & & & & \cmark & & \cmark &  & Fine-grained food recognition &15 &Human\&GPT \\
~~MMVP~\cite{tong2024eyes} &300 &300 & \cmark & & \cmark & & &  &\cmark  &  & Visual perception for CLIP-blind pairs  &8 &Human\&GPT \\
~~HaloQuest~\cite{wang2024haloquest} &7,748 &3,157 & \cmark & \cmark & & \cmark & & &\cmark &  &Hallucination mitigation &16 &GPT \\
~~Visual Riddles~\cite{bittonguetta2024visualriddles} &400 &400 & & \cmark & & \cmark & \cmark & &\cmark & & Visual riddles requiring commonsense &6 &Human\&GPT \\
\midrule
\multicolumn{12}{@{}l@{}}{\textbf{Reasoning beyond Commonsense}} \\
~~ISEKAI~\cite{tai2024link} &-- &1,498  & & \cmark & & & & & & \cmark   &Novel concept learning &5 &N/A \\
~~WHOOPS~\cite{Bitton-Guetta_2023_WHOOPS} &3,374 &500 & & \cmark & & \cmark & & & &\cmark  & Abnormal element explanation &6 &Human\&GPT \\
~~ROME~\cite{zhou2023rome} &-- &1,563 & & \cmark & \cmark & & & & & \cmark 
& Counter-intuitive content interpretation &6 &N/A \\
\midrule
\rowcolor{gray!20}~~\method{} (Ours) &2,700 &900 & \cmark & \cmark & & \cmark & \cmark & \cmark & \cmark & \cmark & Attribute-modified object identification &26 &Human\&GPT \\
\bottomrule[1pt]
\end{NiceTabular}
}
\end{table*}

\section{Related Work}
\label{sec:relatedwork}

\noindent
\textbf{Visual understanding benchmarks for MLLMs.}
They can be roughly categorized into general visual understanding and key capability evaluations (Tab.~\ref{tab:benchmarks}). 
(1) For the former, LLaVA-Bench~\cite{liu2023llava}, MME~\cite{fu2023mme}, MMBench~\cite{liu2025mmbench} and SEED-Bench-1~\cite{li2024seed} focus on basic perception, with SEED-Bench-2~\cite{li2023seedbench2} supporting interleaved image-text input and image\&text output, MMMU~\cite{yue2023mmmu} and MMMU-Pro~\cite{yue2024mmmupro} requiring expert-level domain knowledge.  
(2) For the latter, specialized benchmarks test various abilities, including reasoning on natural adversarial samples (NaturalBench~\cite{naturalbench}), withholding answers (UPD~\cite{miyai2024unsolvable}), fine-grained recognition of food (FoodieQA~\cite{li2024foodieqa}), visual perception for CLIP-blind pairs (MMVP~\cite{tong2024eyes}), hallucination mitigation (HaloQuest~\cite{wang2024haloquest}), and visual riddles using commonsense (Visual Riddles~\cite{bittonguetta2024visualriddles}). 
In contrast, \method{} aims to advance reasoning from commonsense to beyond commonsense.

\noindent
\textbf{Beyond-commonsense benchmarks for MLLMs.} 
Recent benchmarks have emerged to evaluate MLLMs' reasoning beyond commonsense.
ISEKAI~\cite{tai2024link} creates unreal objects by transferring real-world objects into an ``Isekai"  world, testing whether models can learn novel concepts.
WHOOPS~\cite{Bitton-Guetta_2023_WHOOPS} collects ``weird” images with unusual object co-occurrences, assessing if models can explain abnormal elements.
The most relevant work to \method{} is ROME~\cite{zhou2023rome}, which generates counterintuitive images by modifying attributes and positional relations, to evaluates models on interpreting such content.
However, our \method{} differs in several ways: 
(1) \method{} evaluates models' ability to reason from commonsense to beyond commonsense by connecting attribute-modified objects in novel situations to their original counterparts in normal situations.
(2) It provides both original objects and their attribute-modified versions for comprehensive evaluation.
(3) Attribute-modified objects in \method{} are filtered out through human validation to ensure truly novel attributes unseen in the real world.

\begin{figure*}
\centering
\includegraphics[width=1.00\textwidth]{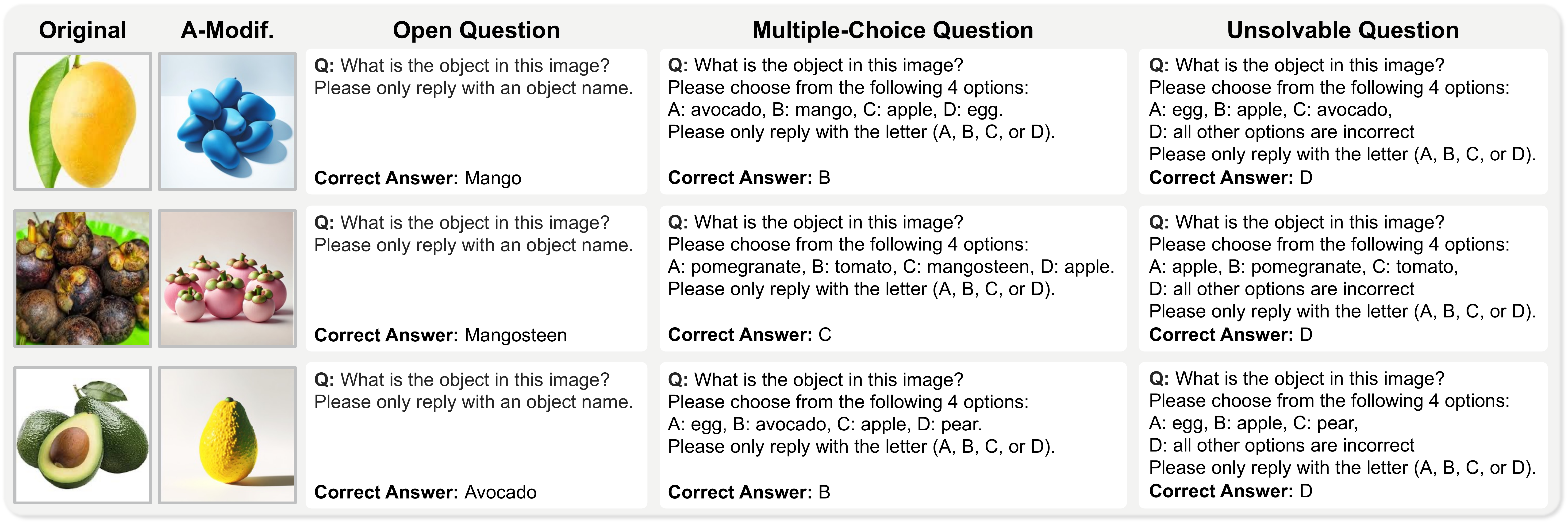}
\caption{Samples from the \method{} benchmark, showcasing original and attribute-modified (A-Modif) objects with three types of question formats: open questions, multiple-choice questions, and unsolvable questions. We additionally circularly shift the choices 3 times in multiple-choice questions and unsolvable questions for extensive evaluation.}
\label{fig_data_samples}
\end{figure*}

\section{\method{} Benchmark}
\label{sec:method}

We here describe how we create \method{}. 
More detail can be found in SM.

\subsection{Selection of Objects}
\label{subsec:selection}
As explained in \cref{sec:intro}, we mainly target fruits to build \method{} since they generally have simple structures and appearances, making it relatively easy to modify their colors.
We select 45 common categories from  Fruit360 dataset~\cite{ukwuoma2022fruit}.
We note that we merge similar categories such as \textit{``passion fruit"} and \textit{``maracuja"} to \textit{``passion fruit"} to lift potential ambiguities for MLLMs and even human.
The full list of categories is in SM.

\subsection{Data Collection}
\label{subsec:collection}
To examine MLLMs' reasoning from commonsense to beyond commonsense, we collect images of both original and attribute-modified objects, with samples displayed in \cref{fig_data_samples}.

\noindent
\textbf{Original objects collection.} 
We collect ten images per category from the internet, resulting in 450 images of original fruits. Each image undergoes a manual check to ensure no additional objects are presented, maintaining dataset clarity.

\noindent
\textbf{Attribute-modified objects collection.}
For each object, we generate ten attribute-modified images using DALL-E~\cite{Aditya2021dalle} with prompts like ``\textit{An image of \{color\} \{object\}}", producing 450 images.
The \textit{color} refers to the novel color provided by GPT-4o~\cite{gpt4v}, verified through manual inspection.
We conduct manual checks and re-generation to ensure 
(1) The modified attributes are distinct from those in the original objects; (2) The attribute-modified objects remain recognizable to humans, meaning their categories are preserved.

\subsection{QA Generation}
\label{subsec:qa}

We created three types of questions to assess MLLMs' reasoning capability in original and attribute-modified objects:

\noindent
\textbf{Open question.}
This type of question asks the model to identify the object in the image without providing any options, relying solely on its own knowledge. 
As the open question examples shown in \cref{fig_data_samples}, the question is ``\textit{What is the object in this image? Please only reply with an object name.}" This format evaluates the model's ability to provide direct, accurate responses.

\noindent
\textbf{Multiple-choice question.}
In these questions, we provide four possible options, including the correct answer and three distractors. The model is expected to select the correct answer by choosing the letter associated with the object name.
As the multiple-choice question examples shown in \cref{fig_data_samples}, the question is like, ``\textit{What is the object in this image? Please choose from the following 4 options: A: $\{distractor_1\}$, B: $\{distractor_2\}$, C: $\{correct\ object\}$, D: $\{distractor_3\}$.}" 
This format evaluates the model’s ability to differentiate between similar or related objects.

\noindent
\textbf{Unsolvable question.}
These questions are designed to test the model's handling of unsolvable problems in the context of visual question answering~\cite{miyai2024unsolvable}. 
In each case, four options are provided, but none of them has the candidate object correctly matching the object in the image. 
The model is expected to select the option indicating that all other choices are incorrect.
As the unsolvable question examples shown in \cref{fig_data_samples}, it is in a format of, ``\textit{What is the object in this image? Please choose from the following 4 options: A: $\{distractor_1\}$, B: $\{distractor_2\}$, C: $\{distractor_3\}$, D: all other options are incorrect.}"
This format evaluates the model’s ability to give a suitable response even when there are no correct objects in the options.

In sum, we have 2,700 (three for each image) questions.

\section{Experimental Settings}
\label{sec:experimental_settings}

\noindent
\textbf{Representative MLLM models.}
We evaluate a total of 26 MLLMs, including \textbf{(1) open-sourced models}: LLaVA-NeXT~\cite{li2024llavanext-strong} (8B, 32B, 72B, 110B), InternVL2~\cite{chen2024far} (1B, 2B, 4B, 8B, 26B, 40B, 76B), 
Cambrian-1~\cite{tong2024cambrian} (3B, 8B, 13B, 34B), 
Yi-VL~\cite{young2024yi} (6B, 34B), 
InstructBLIP~\cite{dai2023instructblip} (7B, 13B), MiniGPT4~\cite{zhu2024minigpt} (7B, 13B), 
MiniGPT-v2~\cite{chen2023minigptv2} (7B); 
and \textbf{(2) commercial models}: GPT-4o~\cite{gpt4v}, GPT-4o-mini~\cite{gpt4v}, 
Gemini-1.5-Flash~\cite{gemini}, Gemini-1.5-Pro~\cite{gemini}.
We used the most up-to-date versions of each model, as of October 2024.

\noindent
\textbf{Circular shifting choice strategy.} 
As suggested in \cite{liu2025mmbench}, MLLMs might prefer to predict a certain choice among multiple choices. We employ a circular shifting choice strategy in multiple-choice and unsolvable questions, resulting in feeding the questions to MLLMs four (number of choices) times with circular shifting choices.

\noindent
\textbf{Model evaluation strategy.} 
For most models, we extract a single word or option directly from the model’s output, or, if needed, parse the output to obtain a single word or option to compare with the ground truth. For models like Cambrian-1~\cite{tong2024cambrian} and MiniGPT4~\cite{zhu2024minigpt}, which often produce sentences, we employ GPT-4o~\cite{gpt4v} to evaluate the alignment between the model’s response and the ground truth. 
A question is considered correctly answered if the model’s output matches the ground truth. We use these results to compute accuracy scores. More details are in SM.

\begin{table*}[tb]
\centering
\caption{Performance in terms of accuracy (\%) scores of 26 representative MLLMs on \method{}, including open-sourced and commercial models, human and random choice baselines. Higher score is better. Results cover original and attribute-modified objects, with performance gaps ($\Delta$) between them for each question type.
\textbf{\colorbox{backblue1}{\makebox(18,5){\vspace{0.15mm}{Bold}}}} marks the best results, and \colorbox{backblue}{\makebox(24,5){\vspace{0.15mm}{original}}} marks the second best.}
\label{tab:overall}
\setlength\tabcolsep{3pt}
\resizebox{\linewidth}{!}{
\begin{NiceTabular}{@{}l|ll|cccc|cc:cc:cc:cc@{}}
\CodeBefore
\rectanglecolor{softgray}{7-1}{10-15} 
\rectanglecolor{softgray}{18-1}{21-15} 
\rectanglecolor{softgray}{24-1}{25-15} 
\rectanglecolor{softgray}{30-1}{31-15} 
\Body
\toprule[1pt]
\multirow{2}{*}{\textbf{Method}} & \multicolumn{2}{c|}{\textbf{Components}} & \multicolumn{4}{c|}{\textbf{Original Objects}} & \multicolumn{8}{c}{\textbf{Attribute-Modified Objects}} \\
\cmidrule(lr){2-3} \cmidrule(lr){4-7} \cmidrule(lr){8-15}
& \multicolumn{1}{c}{Vision Encoder} & \multicolumn{1}{c|}{LLM} 
& \multicolumn{1}{c}{Open} & \multicolumn{1}{c}{Choice} & \multicolumn{1}{c}{Unsol.} & \multicolumn{1}{c|}{Average}
& \multicolumn{1}{c}{Open} & \multicolumn{1}{c}{$\Delta_{\text{O}}$} & \multicolumn{1}{c}{Choice} & \multicolumn{1}{c}{$\Delta_{\text{C}}$} & \multicolumn{1}{c}{Unsol.} & \multicolumn{1}{c}{$\Delta_{\text{U}}$} & \multicolumn{1}{c}{Average} & \multicolumn{1}{c}{$\Delta_{\text{A}}$} \\
\midrule
\multicolumn{15}{l}{\textbf{Baselines}}\\
~~Human & -- & -- & 96.6\std3.6 & 98.7\std0.6 & 98.8\std0.5 & 98.0 &87.3\std7.5  &-9.3  & 88.3\std5.8 & –10.4 & 91.8\std6.6 & -7.0 & 89.1 & -8.9 \\
~~Random choice & -- & -- & -- & 24.4\std2.7 & 27.0\std1.7 & 25.7 & -- & -- & 24.2\std2.5 & –0.2 & 24.1\std2.0 & –2.9 & 24.2 & –1.5 \\
\midrule

\multicolumn{15}{l}{\textbf{Open-Sourced Models}}\\
~~LLaVA-NeXT-8B~\cite{li2024llavanext-strong} & CLIP-ViT-L-14-336 &  LLaMA3$_{\text{8B}}$ & 42.0 & 77.2\std0.8 & 27.1\std13.9 & 48.8 & 24.9 & –17.1 & 43.4\std1.0 & –33.8 & 23.8\std14.2 & –3.3 & 30.7 & –18.1 \\
~~LLaVA-NeXT-Qwen-32B~\cite{li2024llavanext-strong} & SigLIP-ViT-SO-14-384 & Qwen1.5$_{\text{32B}}$ & 58.0 & 93.8\std0.3 & 42.0\std16.5 & 64.6 & 35.8 & –22.2 & 72.0\std1.0 & –21.8 & \second{33.1}\std16.2 & –8.9 & 47.0 & –17.6 \\
~~LLaVA-NeXT-72B~\cite{li2024llavanext-strong} & CLIP-ViT-L-14-336 & Qwen1.5$_{\text{72B}}$ & 47.6 & 84.5\std0.5 & 28.2\std9.7 & 53.4 & 24.2 & –23.4 & 49.4\std0.8 & –35.1 & 21.9\std8.4 & –6.3 & 31.8 & –21.6 \\
~~LLaVA-NeXT-110B~\cite{li2024llavanext-strong} & CLIP-ViT-L-14-336 & Qwen1.5$_{\text{110B}}$ & 51.3 & 85.5\std1.2 & 16.2\std8.3 & 51.0 & 26.9 & –24.4 & 52.2\std0.6 & –33.3 & 9.8\std4.1 & –6.4 & 29.6 & –21.4 \\

~~InternVL2-1B~\cite{chen2024far} & InternViT-300M-448 & Qwen2-Instruct$_{\text{0.5B}}$ & 45.3 & 71.5\std1.0 & 21.6\std10.4 & 46.1 & 24.0 & –21.3 & 50.2\std2.4 & –21.3 & 13.9\std4.6 & –7.7 & 29.4 & –16.7 \\
~~InternVL2-2B~\cite{chen2024far} & InternViT-300M-448 & InternLM2-Chat$_{\text{1.8B}}$ & 49.6 & 68.4\std0.5 & 13.8\std0.4 & 43.9 & 25.6 & –24.0 & 40.4\std0.8 & –28.0 & 8.4\std1.5 & –5.4 & 24.8 & –19.1 \\
~~InternVL2-4B~\cite{chen2024far} & InternViT-300M-448 & Phi3-Mini-128K-Instruct$_{\text{3.8B}}$ & 50.9 & 82.9\std0.7 & 18.6\std6.0 & 50.8 & 23.3 & –27.6 & 52.7\std0.7 & –30.2 & 11.2\std2.5 & –7.4 & 29.1 & –21.7 \\
~~InternVL2-8B~\cite{chen2024far} & InternViT-300M-448 & InternLM2.5-Chat$_{\text{7B}}$ & 50.7 & 84.3\std0.5 & 31.3\std5.2 & 55.4 & 25.8 & –24.9 & 50.2\std1.7 & –34.1 & 22.0\std8.5 & –9.3 & 32.7 & –22.7 \\
~~InternVL2-26B~\cite{chen2024far} & InternViT-6B-448-V1.5 & InternLM2-Chat$_{\text{20B}}$ & \best{79.3} & \best{96.0}\std0.5 & 24.3\std8.1 & 66.5 & \second{53.3} & –26.0 & \best{80.6}\std0.6 & –15.4 & 15.0\std6.8 & –9.3 & 49.6 & –16.9 \\
~~InternVL2-40B~\cite{chen2024far} & InternViT-6B-448-V1.5 & Nous-Hermes2-Yi$_{\text{32B}}$ & 69.6 & \second{95.4}\std0.3 & 44.9\std8.6 & \second{70.0} & 41.3 & –28.3 & \second{73.5}\std1.0 & –21.9 & \best{35.5}\std11.0 & –9.4 & \best{50.1} & –19.9 \\
~~InternVL2-LLaMA3-76B~\cite{chen2024far} & InternViT-6B-448-V1.5 & Hermes2-Theta-LLaMA3$_{\text{70B}}$ & \second{73.3} & 87.0\std0.9 & 35.7\std6.1 & 65.3 & 44.4 & –28.9 & 63.4\std0.5 & –23.6 & 25.0\std6.2 & –10.7 & 44.3 & –21.0 \\

~~Cambrian-1-3B~\cite{tong2024cambrian} & SigLIP-ViT-SO-14 & Phi3$_{\text{3.8B}}$ & 58.9 & 91.8\std0.4 & 44.4\std3.6 & 65.0 & 32.4 & –26.5 & 71.1\std0.7 & –20.7 & 22.7\std4.7 & –21.7 & 42.1 & –22.9 \\
~~Cambrian-1-8B~\cite{tong2024cambrian} & ~~+CLIP-ViT-L-14 & LLaMA3$_{\text{8B}}$ & 62.9 & 91.8\std0.5 & \best{66.8}\std3.7 & \best{73.8} & 36.7 & –26.2 & 67.4\std1.6 & –24.4 & 30.3\std4.5 & –36.5 & 44.8 & –29.0 \\
~~Cambrian-1-13B~\cite{tong2024cambrian} & ~~+DINOv2-g & Vicuna1.5$_{\text{13B}}$ & 59.8 & 88.0\std1.4 & 41.7\std3.4 & 63.2 & 37.1 & –22.7 & 65.5\std1.1 & –22.5 & 23.5\std3.5 & –18.2 & 42.0 & –21.2 \\
~~Cambrian-1-34B~\cite{tong2024cambrian} & ~~+CLIP-ConvNeXT-XXL & Hermes2-Yi$_{\text{34B}}$ & 65.3 & 92.8\std0.1 & \second{48.1}\std5.3 & 68.7 & \best{55.6} & –9.7 & 66.4\std1.1 & –26.4 & 24.9\std2.8 & –23.2 & \second{49.0} & –19.7 \\

~~Yi-VL-6B~\cite{young2024yi} & CLIP-ViT-H-14-224 & Yi-Chat$_{\text{6B}}$ & 57.6 & 87.8\std0.9 & 44.0\std10.4 & 63.1 & 30.4 & –27.2 & 55.5\std1.1 & –32.3 & 24.2\std7.1 & –19.8 & 36.7 & –26.4 \\
~~Yi-VL-34B~\cite{young2024yi} & CLIP-ViT-H-14-224 & Yi-Chat$_{\text{34B}}$ & 53.6 & 91.2\std1.6 & 26.3\std11.2 & 57.0 & 25.8 & –27.8 & 52.7\std1.5 & –38.5 & 19.1\std9.5 & –7.2 & 32.5 & –24.5 \\

~~InstructBLIP-7B~\cite{dai2023instructblip} & EVA01-ViT-G-14-224 & Vicuna$_{\text{7B}}$ & 72.9 & 57.3\std12.6 & 18.2\std29.5 & 49.5 & 23.3 & –49.6 & 44.3\std10.1 & –13.0 & 11.2\std14.5 & –7.0 & 26.3 & –23.2 \\
~~InstructBLIP-13B~\cite{dai2023instructblip} & EVA01-ViT-G-14-224 & Vicuna$_{\text{13B}}$ & 68.0 & 59.1\std14.4 & 10.2\std9.3 & 45.8 & 31.6 & –36.4 & 41.6\std9.3 & –17.5 & 6.5\std7.9 & –3.7 & 26.6 & –19.2 \\
~~MiniGPT4-7B~\cite{zhu2024minigpt} & EVA01-ViT-G-14-224 & Vicuna$_{\text{7B}}$ & 54.9 & 53.8\std1.7 & 14.8\std3.6 & 41.2 & 30.9 & –24.0 & 22.4\std3.5 & –31.4 & 8.7\std4.2 & –6.1 & 20.7 & –20.5 \\
~~MiniGPT4-13B~\cite{zhu2024minigpt} & EVA01-ViT-G-14-224 & Vicuna$_{\text{13B}}$ & 54.9 & 66.4\std2.5 & 5.2\std1.7 & 42.2 & 31.3 & –23.6 & 31.4\std1.3 & –35.0 & 7.3\std2.9 & +2.1 & 23.3 & –18.9 \\
~~MiniGPT-v2-7B~\cite{chen2023minigptv2} & EVA01-ViT-G-14-224 & LLaMA2$_{\text{7B}}$ & 51.6 & 65.1\std4.7 & 21.7\std27.1 & 46.1 & 26.0 & –25.6 & 47.3\std4.3 & –17.8 & 16.7\std23.3 & –5.0 & 30.0 & –16.1 \\

\midrule
\multicolumn{15}{l}{\textbf{Commercial Models}} \\
~~GPT-4o~\cite{gpt4v} & -- & -- & \best{90.9} & \best{99.3}\std0.2 & \second{84.2}\std1.9 & \best{91.5} & 56.2 & –34.7 & 77.6\std1.0 & –21.7 & \second{73.4}\std7.5 & –10.8 & \best{69.1} & –22.4 \\
~~GPT-4o-mini~\cite{gpt4v} & -- & -- & 85.3 & 95.4\std0.4 & \best{87.0}\std1.1 & \second{89.2} & 49.6 & –35.7 & 72.0\std0.9 & –23.4 & \best{77.7}\std2.4 & –9.3 & \second{66.4} & –22.8 \\
~~Gemini-1.5-Flash~\cite{gemini} & -- & -- & 86.2 & 96.1\std1.2 & 57.9\std15.0 & 80.1 & \best{63.3} & –22.9 & \second{79.2}\std0.9 & –16.9 & 29.2\std10.5 & –28.7 & 57.2 & –22.9 \\
~~Gemini-1.5-Pro~\cite{gemini} & -- & -- & \second{86.4} & \second{99.2}\std0.4 & 68.3\std1.0 & 84.6 & \second{62.2} & –24.2 & \best{83.6}\std1.5 & –15.6 & 39.1\std8.8 & –29.2 & 61.6 & –23.0 \\

\bottomrule[1pt]
\end{NiceTabular}
}
\end{table*}

\section{Experimental Results}
\label{sec:experimental_results}

\subsection{Overall Comparison}
\label{subsec:overall}

\cref{tab:overall} presents the evaluations on various MLLMs.
Note that these MLLMs vary in vision encoders and LLMs.

\noindent
\textbf{Human evaluation confirms the quality of \method{}.}
We engage human annotators to answer questions in random order, as detailed in SM.
Human performance serves as the baseline in this comparison, achieving consistently high scores across all question types for both original and attribute-modified objects. 
This high performance reflects the reliability and quality of the constructed dataset, providing a solid foundation for evaluating MLLMs on commonsense and beyond-commonsense reasoning. 

\noindent
\textbf{MLLMs' performance declines on objects with modified attributes.}
Across all models, there is a noticeable performance drop when moving from original to attribute-modified objects. 
This trend highlights the challenge MLLMs face in recognizing objects with modified attributes, requiring beyond-commonsense reasoning. 
For example, Cambrian-1-8B~\cite{tong2024cambrian} achieves an average score of 73.8 on original objects but drops to 44.8 on attribute-modified objects.
Even GPT-4o~\cite{gpt4v}, exhibits a significant decline of 22.4, revealing a general limitation in handling attribute modifications effectively.

When examining the three question types, the performance drop is most pronounced in open and multiple-choice questions when shifting from original to attribute-modified objects.
For unsolvable questions, both open-sourced and certain commercial models, such as Gemini~\cite{gemini}, perform poorly on original and attribute-modified objects with high standard deviations. This variability indicates instability and inconsistency in MLLMs' responses when choices are circularly shifted in unsolvable questions.

\noindent
\textbf{Open-sourced MLLMs underperform compared to commercial models, especially on attribute-modified objects.}
Commercial models, such as GPT-4o~\cite{gpt4v} and GPT-4o-mini~\cite{gpt4v}, lead in performance across both original and attribute-modified objects, with GPT-4o reaching a top average of 91.5 on original objects and 69.1 on attribute-modified objects. 
In contrast, open-sourced MLLMs, like Cambrian-1-8B~\cite{tong2024cambrian} and InternVL2-40B~\cite{chen2024far}, generally score lower, with averages of 73.8 on original objects and 50.1 on attribute-modified objects, respectively.

Furthermore, the performance gap between commercial and open-sourced MLLMs widens further on attribute-modified objects. 
For example, the gap between GPT-4o~\cite{gpt4v} and Cambrian-1-8B~\cite{tong2024cambrian}  is 17.7 on original objects, while it becomes 24.3 on attribute-modified objects.

Overall, these results indicate current MLLMs, even GPT-4o~\cite{gpt4v}, need substantial improvements to handle beyond-commonsense reasoning effectively.
Therefore, we explore performance decline within the MLLM architecture in \cref{subsec:model}, model preferences for certain objects within \method{} in \cref{subsec:data}, and samples outside \method{} in \cref{subsec:other}, to offer deeper insights into this problem.

\subsection{Probing Performance Drop within Model Itself}
\label{subsec:model}

In this section, we examine the reasons behind the performance drop observed when open-sourced MLLMs encounter attribute-modified objects from the aspects of model components and model sizes.

\begin{figure}
\centering
\includegraphics[width=0.475\textwidth]{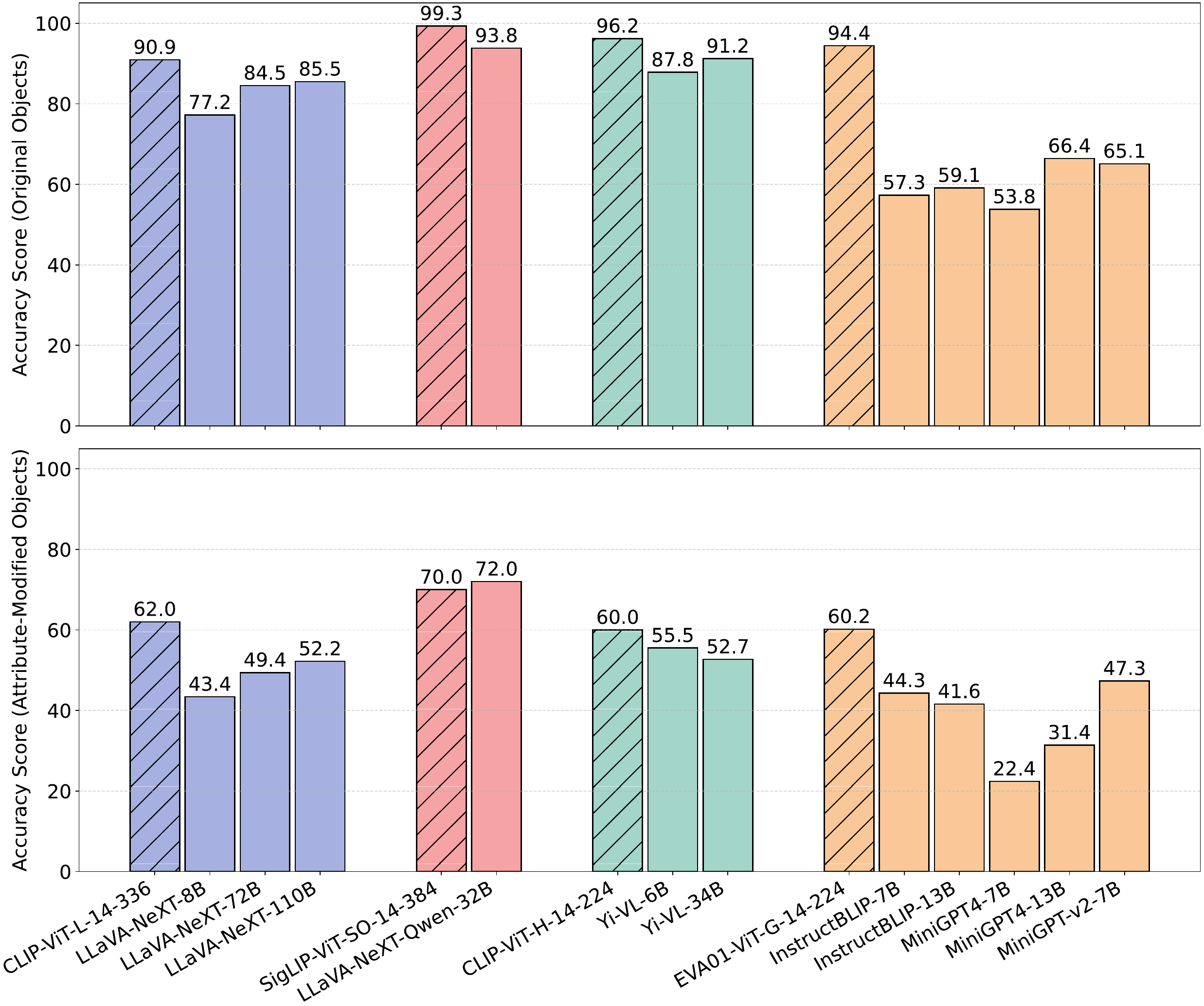}
\vspace*{-0.8\baselineskip}
\caption{Model accuracy scores across vision encoders for multiple-choice questions on original objects (upper) and attribute-modified objects (lower). MLLMs using the same vision encoder are shown in the same color, while vision encoders themselves are distinguished by diagonal line patterns. Stronger encoders generally improve MLLM performance, yet most MLLMs perform worse than their encoders.}
\label{fig_visen}
\vspace*{-1.0\baselineskip}
\end{figure}

\noindent
\textit{\textbf{Question 1: How does the pre-trained vision encoder in MLLMs contribute to the performance gap?}}

\noindent
We compare the model performance in original objects (\cref{fig_visen}, upper) and attribute-modified objects (\cref{fig_visen}, lower), focusing on the different vision encoders used in MLLM models, including CLIP-ViT-L-14-336~\cite{radford2021clip}, SigLIP-ViT-SO-14-384~\cite{zhai2023siglip}, CLIP-ViT-H-14-224~\cite{radford2021clip}, and EVA01-ViT-G-14-224~\cite{EVA-CLIP}.
Note here the results for the vision encoder come from the pre-trained vision encoder. 
To get accuracy scores for vision encoders, we compare the image embeddings from the vision encoder, with the text embeddings generated by corresponding text encoder for object names in the multiple-choice questions.
So, for fair comparison, the results of MLLMs presented here are from the multiple-choice accuracy score.

As shown in \cref{fig_visen} upper, if pre-trained vision encoders perform better, MLLMs with the same vision encoder also generally perform well on original objects.
Specifically, models with encoders such as SigLIP-ViT-SO-14-384~\cite{zhai2023siglip} and CLIP-ViT-H-14-224~\cite{radford2021clip} demonstrate high accuracy across different MLLMs. 
However, a surprising finding is that, across most models, a single pre-trained vision encoder consistently outperforms the MLLMs that use it. This discrepancy suggests that the current MLLMs' vision-language alignment might be a limiting factor, as the projector and LLM fail to fully leverage the robust embeddings provided by high-performance pre-trained vision encoders.

A similar pattern emerges in the results on attribute-modified objects, as seen in \cref{fig_visen} lower. 
Interestingly, LLaVA-NeXT-Qwen-32B~\cite{li2024llavanext-strong}, utilizing the SigLIP-ViT-SO-14-384~\cite{zhai2023siglip} vision encoder, demonstrates a notable exception by achieving a slightly higher accuracy on attribute-modified objects than the standalone performance of its vision encoder. 
This suggests that being equipped with enough strong vision encoders may help mitigate some of the performance losses on modified attributes.

\begin{tcolorbox}[colframe=black, colback=gray!20, arc=2pt, left=0.2cm,right=0.2cm, top=2pt, bottom=2pt, boxrule=1pt]
\small
\textbf{Answer 1:} While advanced vision encoders contribute to improved recognition in original and modified objects, the full potential of these encoders is not fully realized in current MLLM models. Addressing this performance gap will likely require innovations in integrating vision and language components more effectively.
\end{tcolorbox}

\noindent
\textit{\textbf{Question 2: How does MLLM size affect the gap?}}

\begin{figure}
\centering
\includegraphics[width=0.45\textwidth]{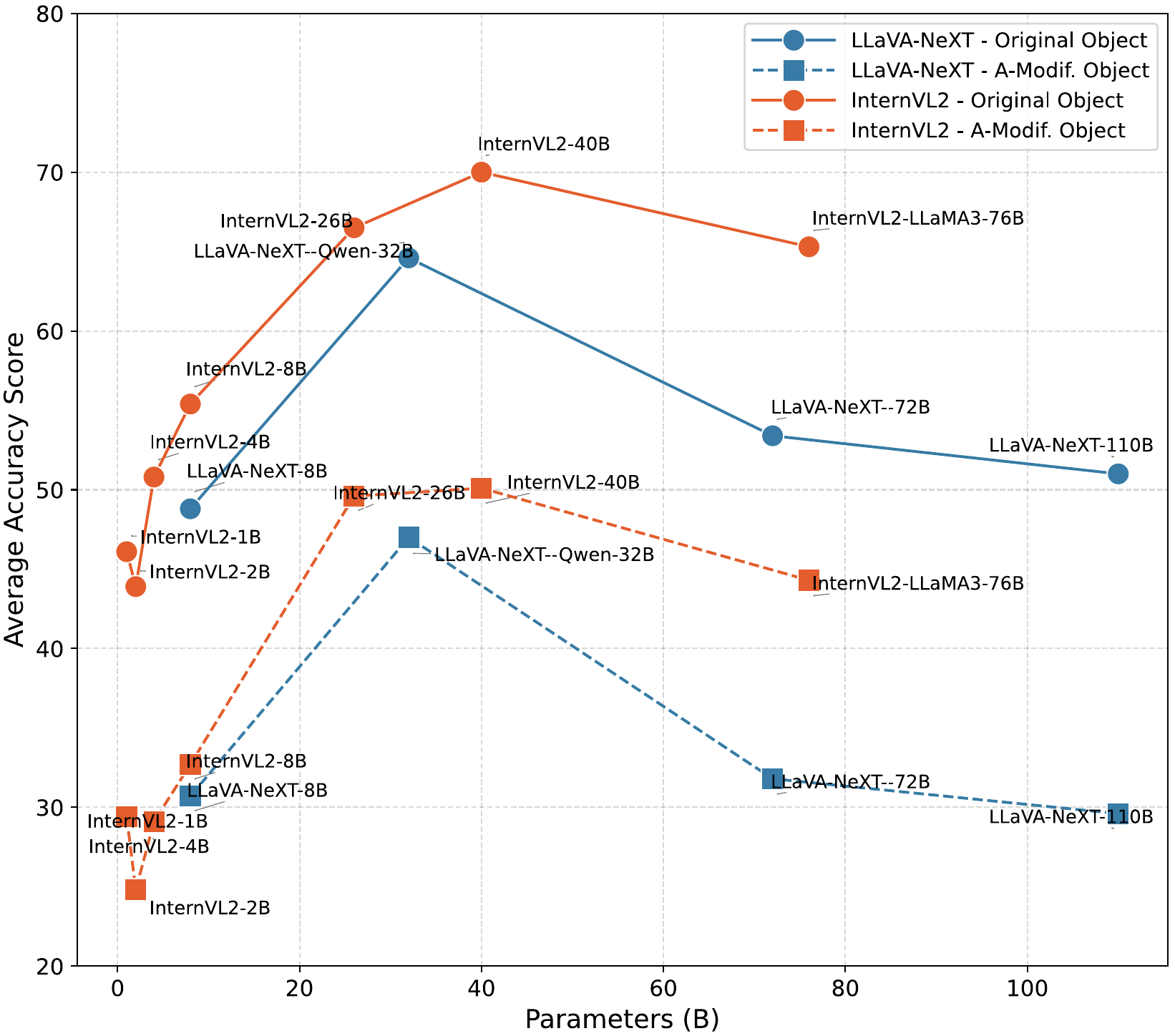}
\vspace*{-0.5\baselineskip}
\caption{Average accuracy scores of various MLLMs (LLaVA-NeXT and InternVL2 series) for all question types across different model sizes on original and attribute-modified objects. The performance does not consistently improve as the model size increases.}
\label{fig_model_size}
\vspace*{-1.0\baselineskip}
\end{figure}

\noindent 
We analyze average scores of MLLMs for all question types with varying model sizes, including LLaVA-NeXT~\cite{li2024llavanext-strong} (8B to 110B) and InternVL2~\cite{chen2024far} (1B to 76B), on both original and attribute-modified objects, as shown in ~\cref{fig_model_size}. 

For both original and attribute-modified objects, increasing the model size generally leads to improved performance initially. 
Smaller models such as InternVL2-1B and LLaVA-NeXT-8B show relatively low accuracy on both object types, while accuracy improves with models like InternVL2-26B and LLaVA-NeXT-Qwen-32B, suggesting MLLMs benefit from more parameters.

However, as model size continues to increase, we observe a decline in accuracy, particularly on attribute-modified objects.
For example, from LLaVA-NeXT-Qwen-32B to LLaVA-NeXT-72B, performance decreases by 15.2 on attribute-modified objects and by 11.2 on original objects.
This indicate that, beyond a certain size, additional parameters may add complexity that hinders the model's adaptation to modified attributes.

\begin{tcolorbox}[colframe=black, colback=gray!20, arc=2pt, left=0.2cm,right=0.2cm, top=2pt, bottom=2pt, boxrule=1pt]
\small
\textbf{Answer 2:} Simply scaling up model size does not guarantee better handling of attribute modifications. Future research should consider optimizing model architectures and training techniques specifically to address the challenges posed by attribute-modified objects, rather than relying solely on increased model size. 
\end{tcolorbox}

\begin{figure}
\centering
\includegraphics[width=0.98\linewidth]{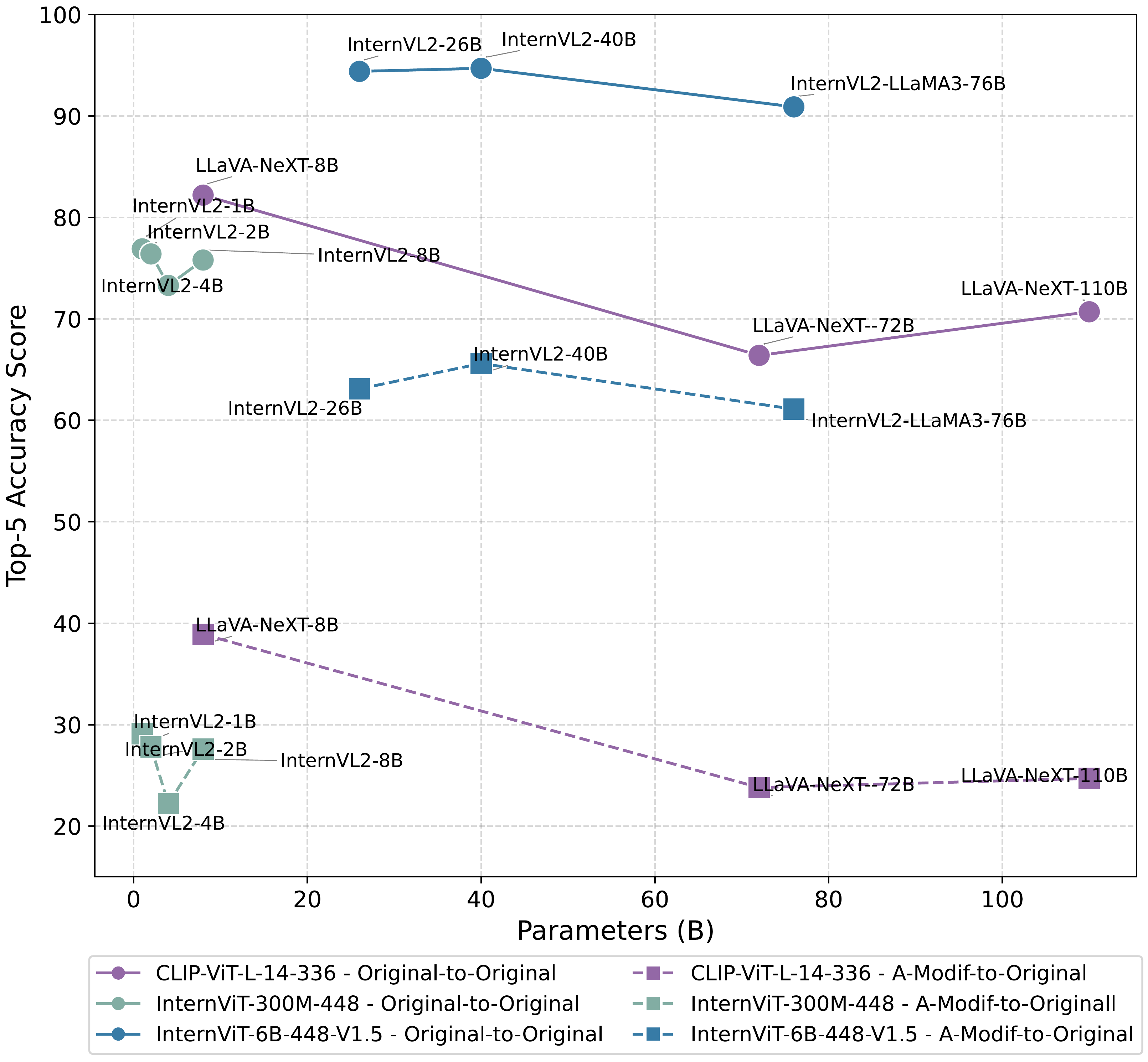}
\caption{Top-5 accuracy of Original-to-Original and Attribute-modified-to-Original object matching, using image representations obtained from MLLMs' trained vision encoders. Fine-tuning with larger LLMs may weaken the vision encoders.}
\label{fig:image_retrieval}
\end{figure}

\noindent
\textit{\textbf{Question 3: Why does scaling up MLLM size not guarantee better performance?}}

\noindent
To answer this question, we separate vision encoders and LLMs in MLLMs to analyze how scaling up each component affects performance. 
Unlike \textit{\textbf{Question 1}}, which focuses on pretrained vision encoders, we separate the vision encoder from the whole MLLMs after fine-tuning; then evaluate its image representation.
We employ image-to-image matching for this experiment including Original-to-Original (O2O) and Attribute-modified-to-Original (A2O) matching.
For each query, a matching is considered correct if one of the top five similar images belongs to the same category (details in SM).

~\cref{fig:image_retrieval} shows that fine-tuning the same pre-trained vision encoder with different LLMs in MLLMs yields varied performance. 
Vision encoders fine-tuned on InternViT-300M-448 perform similarly, possibly because their small scale MLLM limits further improvement. 
In contrast, larger MLLMs like LLaVA-NeXT-72B and LLaVA-NeXT-110B, perform worse than LLaVA-NeXT-8B.
A similar, albeit less pronounced, trend is observed for InternVL2-26B, InternVL2-40B, and InternVL2-LLaMA3-76B, suggesting that fine-tuning with larger LLMs may weaken the vision encoder’s performance.
As noted in \textit{\textbf{Question 2}}, LLaVA-NeXT-8B achieves comparable performance to LLaVA-NeXT-72B and LLaVA-NeXT-110B on attribute-modified objects, despite significantly larger scales of the latter two.

\begin{tcolorbox}[colframe=black, colback=gray!20, arc=2pt, left=0.2cm,right=0.2cm, top=2pt, bottom=2pt, boxrule=1pt]
\small
\textbf{Answer 3:} Stronger vision encoders improve MLLM performance, but larger LLMs often weaken vision encoders during fine-tuning. 
Preserving vision encoder strength is essential, especially when fine-tuning with larger LLMs.
\end{tcolorbox}

\begin{figure*}
\centering
\includegraphics[width=1.00\textwidth]{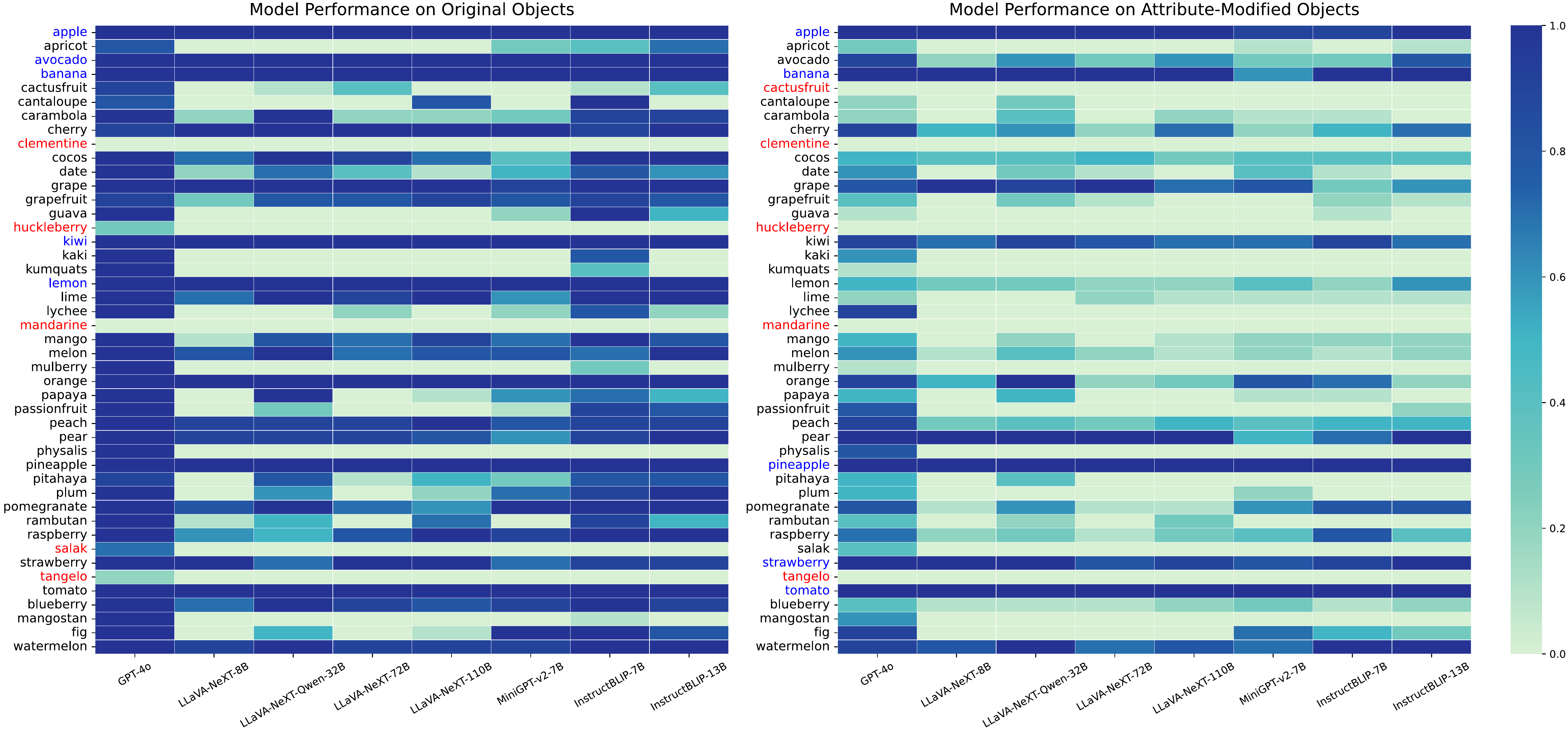}
\vspace*{-1.0\baselineskip}
\caption{Comparison of model performance on different fruits in original objects (left) and attribute-modified objects (right) for open questions. Each heatmap illustrates the accuracy of various models (x-axis) when applied to different fruit types (y-axis), with \colorbox{darkblue}{\makebox(18,5){\vspace{0.15mm}{}}} indicating higher performance and \colorbox{lightgreen}{\makebox(18,5){\vspace{0.15mm}{}}} indicating lower performance. Top-5 object names which all MLLMs prefer most and least are highlighted in \textcolor{blue}{blue} and \textcolor{red}{red}, respectively.}
\label{fig_model_accuracy_objects}
\vspace*{-1.0\baselineskip}	
\end{figure*}

\subsection{Exploring Model Preferences in \method{}}
\label{subsec:data}

\noindent
\textit{\textbf{Question 4: Do MLLMs prefer certain objects?}}

\noindent
In Fig~\ref{fig_model_accuracy_objects}, we compare model performance on both original and attribute-modified objects, grouped by object categories. This analysis provides insights into how different MLLMs perform across objects and if certain categories are easier or more challenging for the models to recognize.

Overall, for original objects, MLLMs perform best on common and easily recognizable objects, such as \textit{``apple"}, \textit{``avocado"}, and \textit{``banana"}.
These objects, due to their familiarity and distinctive features, seem to pose fewer challenges for models. 
However, performance declines significantly for less common or visually complex objects, such as \textit{``clementine"}, \textit{``huckleberry"}, and \textit{``salak"}.
The degradation is amplified when applied to objects with modified attributes, suggesting rare or less familiar objects are especially challenging for MLLMs to recognize accurately when presented with modified attributes.

\begin{tcolorbox}[colframe=black, colback=gray!20, arc=2pt, left=0.2cm,right=0.2cm, top=2pt, bottom=2pt, boxrule=1pt]
\small
\textbf{Answer 4:} MLLMs show a preference for certain objects, particularly those that are common and easily recognizable. In contrast, they struggle with less common objects, especially when objects are presented with modified attributes. This suggests that while MLLMs are effective at recognizing frequently encountered items, they face significant challenges in generalizing to rare or atypical objects, particularly when those objects have non-standard appearances. 
\end{tcolorbox}

\subsection{Investigating outside \method{}}
\label{subsec:other}

In this section, we briefly discuss the results on our \method{}-extension which consists of shape-modified fruits and color-modified animals.
Refer to SM for more details.

\noindent
\textit{\textbf{Question 5: How do MLLMs perform on objects with other modified attributes rather than colors?}}

\noindent In addition to color modifications, we explore the performance of MLLMs on objects with other attribute changes by generating shape-modified images. Using DALL-E~\cite{Aditya2021dalle}, we generate images with prompts in the format, ``\textit{An image of \{shape\} \{object\}}", which resulted in a collection of 450 images with modified shapes, ultimately yielding 260 shape-modified images after selection.

The results of MLLMs on objects with modified shapes are provided in SM, showing a performance pattern similar to that observed on color-modified objects (Tab.~\ref{tab:overall}).

\begin{tcolorbox}[colframe=black, colback=gray!20, arc=2pt, left=0.2cm,right=0.2cm, top=2pt, bottom=2pt, boxrule=1pt]
\small
\textbf{Answer 5:} MLLMs face comparable challenges when recognizing objects with non-standard shapes as they do with color alterations. This consistency suggests that the difficulties MLLMs encounter are not solely tied to specific types of attribute modifications but rather to any deviation from typical object appearances.  
\end{tcolorbox}

\noindent
\textit{\textbf{Question 6: How do MLLMs perform on other objects rather than fruits?}}

\noindent
To extend our analysis beyond fruits, we select 42 animal categories from ImageNet~\cite{deng2009imagenet}.
The generation process for attribute-modified images follows a similar approach, using prompts in the format, ``\textit{An image of \{animal\} with \{color\} skin, replacing its usual skin}".
After filtering, we get 218 images of animals with modified attributes.

Our SM shows the results. Consistent with prior findings, MLLMs show a decline in recognition performance on objects with modified attributes. Additionally, open-sourced MLLMs continue to underperform compared to commercial models like GPT-4o and Gemini-1.5-Pro, both on original and attribute-modified animal objects.

\begin{tcolorbox}[colframe=black, colback=gray!20, arc=2pt, left=0.2cm,right=0.2cm, top=2pt, bottom=2pt, boxrule=1pt]
\small
\textbf{Answer 6:} MLLMs exhibit similar performance trends on non-fruit objects as on fruits. This further demonstrates the limitations of MLLMs in generalizing to attribute-modified representations across various object types, not just fruits.
\end{tcolorbox}
\section{Discussion and Limitations}
\label{sec:discussion}

\noindent
First, a primary limitation in using generated data for model evaluation is the lack of diversity~\cite{Fan_2024_scaling},  which can influence the visual characteristics of generated images. We will explore methods to mitigate these biases in generated data, such as use well-designed prompts to increase the diversity. 

\noindent
Second, while this study focuses primarily on object recognition as a component of semantic reasoning in visual understanding, effective object understanding requires a broader range of skills, including spatial reasoning and contextual interpretation. Future research should aim to evaluate MLLMs across a wider range of visual tasks to better understand their limitations in consistently interpreting both original and attribute-modified objects.

\section{Conclusion}
\label{sec:conclusion}

To answer \textit{``Could MLLMs identify objects when they are modified by beyond-commonsense attributes?"}, we propose \method{} to evaluate 26 MLLMs’ performance on both original and attribute-modified objects, 
revealing that while MLLMs excel with common objects, their accuracy drops notably for objects with modified attributes. 
Open-sourced models, in particular, lag behind commercial models like GPT-4o. 
Key factors contributing to this gap include vision encoder choice, training processes, and model size, with MLLMs struggling to fully utilize strong vision encoders.
We believe that using \method{} as a testbed could lay out potential roadmaps to enhance versatile and resilient MLLMs.

{
    \small
    \bibliographystyle{ieeenat_fullname}
    \bibliography{main}
}


\newpage
\setcounter{section}{0}
\setcounter{figure}{0}
\setcounter{table}{0}

\renewcommand\thesection{\Alph{section}}
\renewcommand\thefigure{\Alph{figure}} 
\renewcommand\thetable{\Alph{table}}

\definecolor{cvprblue}{rgb}{0.21,0.49,0.74}
\definecolor{Gray}{gray}{0.9}
\definecolor{demphcolor}{RGB}{144,144,144}
\definecolor{backblue}{RGB}{221,239,251}
\definecolor{backblue1}{RGB}{186,216,242}
\definecolor{backred}{RGB}{244,199,204}

\definecolor{original}{RGB}{153, 153, 255}
\definecolor{amodif}{RGB}{255, 153, 153}
\definecolor{softgray}{rgb}{0.9, 0.9, 0.9}

\def\confName{CVPR}
\def\confYear{2025}

\twocolumn[
\centering
\section*{
\Large
Supplementary Material for \\ \methodemojititle: Can Multimodal LLMs Identify Attribute-Modified Objects?}
\vspace{0.5cm}
]

This supplementary material complements our paper with the following sections: 
\begin{itemize}[leftmargin=*]

\item We discuss further related work on multimodal large language models (MLLMs) (see \cref{sup_sec_related_work}), which were not covered in the main paper Sec.~2.

\item A detailed description of the methodology used to construct our proposed \method{} benchmark is provided (see \cref{sup_sec_nemo}).

\item We elaborate on the experimental setup and methods used to assess model performance (see \cref{sup_sec_eval}).

\item We provide analysis on failure cases observed in the results of MLLMs on \method{} (see \cref{sup_sec_failure}).

\item We include additional evaluations of MLLMs on \method{}-extension including shape-modified fruits and color-modified animals, as well as \method{} with synthetic original objects and different open question prompts (see \cref{sup_sec_analysis}).

\item We discuss further considerations and observations related to this work (see \cref{sup_sec_discuss}).

\end{itemize}

\section{More Related Work}
\label{sup_sec_related_work}

\noindent
\textbf{Multimodal large language models.} 
Recent open-sourced MLLMs, such as LLaVA-NeXT~\cite{li2024llavanext-strong},  Yi-VL~\cite{young2024yi}, InternVL2~\cite{chen2024far}, Cambrian-1~\cite{tong2024cambrian}, and xGen-MM-base-4B (BLIP-3)~\cite{xue2024blip3} have been demonstrated strong capabilities in various visual-centric tasks. These models typically consist of three components: (1) a visual encoder, like CLIP vision encoder~\cite{radford2021clip, Fang_2023_eva, zhai2023siglip}, which processes visual inputs; 
(2) a pre-trained large language model, such as Vicuna~\cite{vicuna2023}, LLaMA~\cite{touvron2023llama}, Qwen~\cite{qwen2023}, and InternLM~\cite{cai2024internlm2}, to handle multimodal embeddings and perform reasoning; 
and (3) a visual-language projector, such as cross-attention
layers~\cite{alayrac2022flamingo}, MLPs~\cite{liu2023llava,li2024llavanext-strong}, a single linear layer~\cite{zhu2024minigpt, chen2023minigptv2}, and Q-Former~\cite{li2023blip, dai2023instructblip}, to align the visual and language modalities. Given the diversity of MLLMs, evaluating them effectively is a critical challenge.

\section{More Details about \method{}}
\label{sup_sec_nemo}

\subsection{Full List of Categories}
The full list of 45 categories included in our \method{}  is presented in the \textit{``Object"} column of \cref{suptab_fruit_colors}.

\subsection{Original Data Source}

\noindent
\textbf{Image links of original data.}
The images of original objects are collected from the Internet (\cref{sup_fig_original}). Due to copyright considerations, we will only provide the image links for original objects in our public \method{}.

\subsection{Attribute-Modified Data Construction}

\noindent
\textbf{Full list of modified attributes.}
The modified color refers to the novel color provided by GPT-4o~\cite{gpt4v} (\cref{suptab_fruit_colors}). These modifications are manually verified to ensure quality and accuracy.

\begin{table*}[tb]
\centering
\caption{All objects and the modified colors for each object in \method{}.}
\label{suptab_fruit_colors}
\setlength\tabcolsep{3pt}
\resizebox{\linewidth}{!}{
\begin{NiceTabular}{@{}l|llllllllll@{}}
\toprule[1pt]
\textbf{Object} & \textbf{Color$_1$} & \textbf{Color$_2$} & \textbf{Color$_3$} & \textbf{Color$_4$} & \textbf{Color$_5$} & \textbf{Color$_6$} & \textbf{Color$_7$} & \textbf{Color$_8$} & \textbf{Color$_9$} & \textbf{Color$_{10}$} \\
\midrule
apple & bright yellow & vibrant orange & rich purple & deep cyan & light green & sky blue & soft green & soft tangerine & light pink & blush pink \\
apricot & bright yellow & deep purple & vibrant green & lemon yellow & crimson red & soft pink & warm brown & deep cyan & sky blue & bright blue \\
avocado & soft pink & bright yellow & bright red & vibrant orange & light green & crimson red & golden & rich purple & bright cyan & blush pink \\
banana & bright red & deep purple & vibrant orange & light green & soft pink & vibrant pink & light green & deep red & rich blue & vivid red \\
blueberry & soft pink & bright yellow & light purple & deep red & vibrant orange & light green & warm brown & golden & vivid pink & bright red \\
cactusfruit & vibrant pink & deep red & bright red & warm brown & soft pink & lemon yellow & rich purple & bright red & bright yellow & vivid red \\
cantaloupe & bright blue & deep purple & golden & bright red & soft pink & light green & vivid red & vibrant orange & vivid pink & vibrant red \\
carambola & bright red & soft pink & bright blue & deep purple & deep orange & deep red & warm brown & sky blue & deep cyan & deep green \\
cherry & bright yellow & vibrant green & soft pink & vibrant orange & deep purple & vibrant green & light yellow & warm brown & bright cyan & deep blue \\
clementine & bright purple & deep purple & bright red & soft pink & vibrant green & deep red & bright blue & bright red & light green & vivid pink \\
cocos & soft pink & light green & bright red & vibrant orange & bright yellow & golden & rich purple & deep red & vibrant green & blush pink \\
date & bright yellow & soft pink & deep purple & bright red & light green & vibrant orange & warm brown & deep red & vivid pink & golden \\
fig & golden & soft pink & bright yellow & rich purple & deep red & vibrant orange & light blue & warm brown & vivid pink & bright red \\
grape & soft pink & rich cyan & deep red & vibrant orange & light cyan & warm brown & vivid pink & golden & bright blue & bright yellow \\
grapefruit & rich purple & bright yellow & soft pink & deep red & vibrant orange & light green & warm brown & golden & vivid pink & bright red \\
guava & bright yellow & rich purple & soft pink & deep red & vibrant orange & light green & warm brown & golden & vivid pink & bright red \\
huckleberry & soft pink & bright yellow & deep red & rich purple & vibrant orange & light green & vibrant blue & golden & vivid pink & bright red \\
kaki & soft pink & bright blue & rich purple & deep red & light green & vibrant cyan & warm brown & golden & bright red & vivid pink \\
kiwi & bright yellow & soft pink & rich purple & deep red & bright blue & light purple & deep cyan & bright blue & vivid pink & bright red \\
kumquats & soft pink & rich purple & deep red & vibrant blue & light green & warm brown & bright purple & vivid pink & bright red & bright blue \\
lemon & soft pink & rich purple & bright purple & deep red & vibrant orange & light green & warm brown & golden & vivid pink & bright red \\
lime & soft pink & bright blue & rich purple & bright cyan & vibrant blue & light pink & light red & bright red & vivid pink & bright red \\
lychee & bright yellow & rich purple & golden & soft pink & deep red & vibrant orange & light green & sky blue & bright red & vivid pink \\
mandarine & rich purple & bright blue & soft pink & deep red & vibrant cyan & light green & warm purple & sky blue & vivid pink & bright red \\
mango & soft pink & deep red & rich purple & bright blue & vibrant orange & light green & warm brown & golden & vivid pink & bright red \\
mangostan & soft pink & bright yellow & rich blue & deep red & vibrant orange & light green & bright cyan & golden & vivid pink & bright red \\
melon & bright purple & soft pink & black & deep red & bright cyan & light green & bright magenta & golden & vivid pink & bright red \\
mulberry & soft pink & bright yellow & rich purple & light pink & vibrant orange & light green & warm brown & golden & turquoise & vivid pink \\
orange & soft pink & light purple & deep red & vibrant turquoise & light green & light cyan & violet & vivid pink & sky blue & bright magenta \\
papaya & soft pink & bright yellow & rich purple & deep red & vibrant turquoise & light green & light purple & golden & vivid pink & bright red \\
passionfruit & bright yellow & soft pink & rich purple & bright cyan & vibrant orange & sky blue & golden & vivid pink & light green & bright red \\
peach & vibrant turquoise & soft purple & bright yellow & rich purple & deep red & vibrant orange & light green & sky blue & golden & bright cyan \\
pear & bright yellow & soft pink & rich purple & deep red & vibrant orange & light green & warm brown & golden & vivid pink & bright red \\
physalis & soft pink & bright yellow & rich purple & deep red & vibrant orange & light green & warm brown & golden & sky blue & bright red \\
pineapple & soft pink & sky blue & rich purple & deep red & vibrant orange & light green & vivid pink & warm brown & golden & bright red \\
pitahaya & bright yellow & sky blue & rich purple & soft purple & vibrant orange & light green & vibrant turquoise & golden & vivid green & bright cyan \\
plum & soft pink & bright yellow & rich purple & deep red & light green & warm brown & golden & vivid pink & bright cyan & vibrant orange \\
pomegranate & soft pink & bright yellow & deep cyan & vibrant orange & rich purple & light green & warm brown & golden & bright blue & vivid pink \\
rambutan & soft pink & bright yellow & rich purple & sky blue & warm brown & vibrant orange & light green & golden & vivid pink & bright blue \\
raspberry & golden & bright yellow & soft pink & rich purple & deep blue & vibrant orange & light green & warm brown & bright cyan & bright purple \\
salak & soft pink & bright yellow & rich purple & deep red & vibrant orange & light green & warm brown & golden & vivid pink & bright red \\
strawberry & bright yellow & soft pink & rich purple & deep cyan & vibrant orange & light green & warm brown & golden & vivid pink & bright blue \\
tangelo & soft pink & bright blue & rich purple & deep red & vibrant cyan & turquoise & light green & warm purple & vivid pink & bright red \\
tomato & soft pink & bright yellow & rich purple & deep cyan & vibrant orange & light green & warm brown & golden & vivid pink & bright blue \\
watermelon & soft pink & bright yellow & rich purple & deep red & vibrant orange & light blue & warm brown & golden & vivid pink & bright red \\
\bottomrule[1pt]
\end{NiceTabular}
}
\end{table*}

\noindent
\textbf{Image generation.}
We utilize the OpenAI’s DALL-E3~\cite{Aditya2021dalle} API with the model name of ``dall-e-3", size of ``1024$\times$1024", quality of ``standard", to generate attribute-modified images.

\subsection{Questions Construction}

\noindent
\textbf{Multiple choices.}
To construct distractor choices for the multiple-choice questions and unsolvable questions, We collect misclassified objects from the open questions's responses by GPT-4o~\cite{gpt4v}, LLaVA-NeXT-8B~\cite{li2024llavanext-strong}, InstructBLIP-7B~\cite{dai2023instructblip}
, InstructBLIP-13B~\cite{dai2023instructblip}, and MiniGPT-v2-7B~\cite{chen2023minigptv2}. 
We select Top-3  most frequent misclassified objects as the distractor options.
For the objects with less than 3 misclassified objects, we mamually add the distractor options after asking GPT-4o~\cite{gpt4v} to give some suggestions by prompts like \textit{``Could you provide a list of objects that are often mistaken for \{object\} due to their similar appearance, color, or shape?"}
The distractor options for each object are detailed in \cref{suptab_distractor}.

\begin{table}[tb]
\centering
\caption{Distractor options for each object in the  multiple-choice and unsolvable questions of \method{}.}
\label{suptab_distractor}
\resizebox{0.98\linewidth}{!}{
\begin{NiceTabular}{@{}l|lll@{}}
\toprule[1pt]
\textbf{Object} & \textbf{Distractor$_1$} & \textbf{Distractor$_2$}  &\textbf{Distractor$_3$} \\
\midrule
apple & orange & tomato & peach \\
apricot & ball & plum & peach \\
avocado & egg & pear & apple \\
banana & apple & orange & papaya \\
cactusfruit & tomato & berry & egg \\
cantaloupe & apple & watermelon & orange \\
carambola & star & flower & starfish \\
cherry & apple & bowl & orange \\
clementine & orange & apple & lemon \\
cocos & ball & onion & apple \\
date & pear & pomegranate & apple \\
grape & raspberry & ball & apple \\
grapefruit & orange & apple & ball \\
guava & apple & pumpkin & lemon \\
huckleberry & pomegranate & blueberry & bunch \\
kiwi & apple & tomato & orange \\
kaki & tomato & apple & pumpkin \\
kumquats & tomato & egg & orange \\
lemon & orange & ball & lime \\
lime & lemon & orange & ball \\
lychee & apple & strawberry & raspberry \\
mandarine & apple & tomato & orange \\
mango & egg & apple & avocado \\
melon & ball & pumpkin & apple \\
mulberry & raspberry & berry & grape \\
orange & apple & ball & lemon \\
papaya & banana & pumpkin & pear \\
passionfruit & pomegranate & orange & ball \\
peach & apple & quince & tomato \\
pear & apple & orange & strawberry \\
physalis & apple & tomato & berry \\
pineapple & apple & plum & strawberry \\
pitahaya & cactus & pineapple & artichoke \\
plum & apple & orange & apricot \\
pomegranate & apple & pumpkin & pear \\
rambutan & ball & cactu & mushroom \\
raspberry & apple & grape & blackberry \\
salak & orange & egg & pineapple \\
strawberry & blueberry & apple & orange \\
tangelo & orange & apple & lemon \\
tomato & apple & kaki & cherry \\
blueberry & apple & orange & ball \\
mangostan & apple & pomegranate & tomato \\
fig & onion & tomato & pomegranate \\
watermelon & pumpkin & ball & apple \\
\bottomrule[1pt]
\end{NiceTabular}
}
\end{table}

\subsection{Data Examples}
\begin{figure}
\centering
\includegraphics[width=0.99\linewidth]{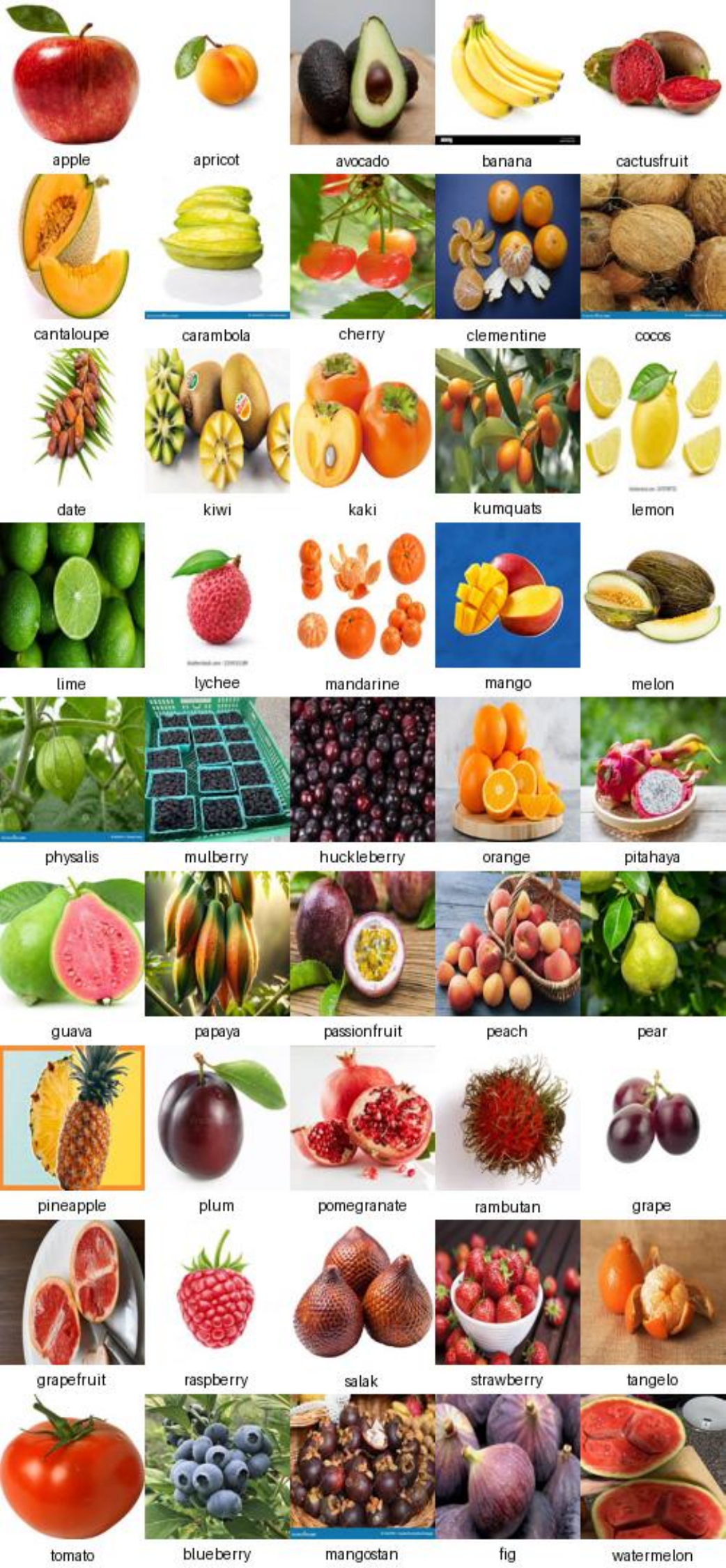}
\caption{Sample for each original object in \method{}.}
\label{sup_fig_original}
\end{figure}

\begin{figure}
\centering
\includegraphics[width=0.99\linewidth]{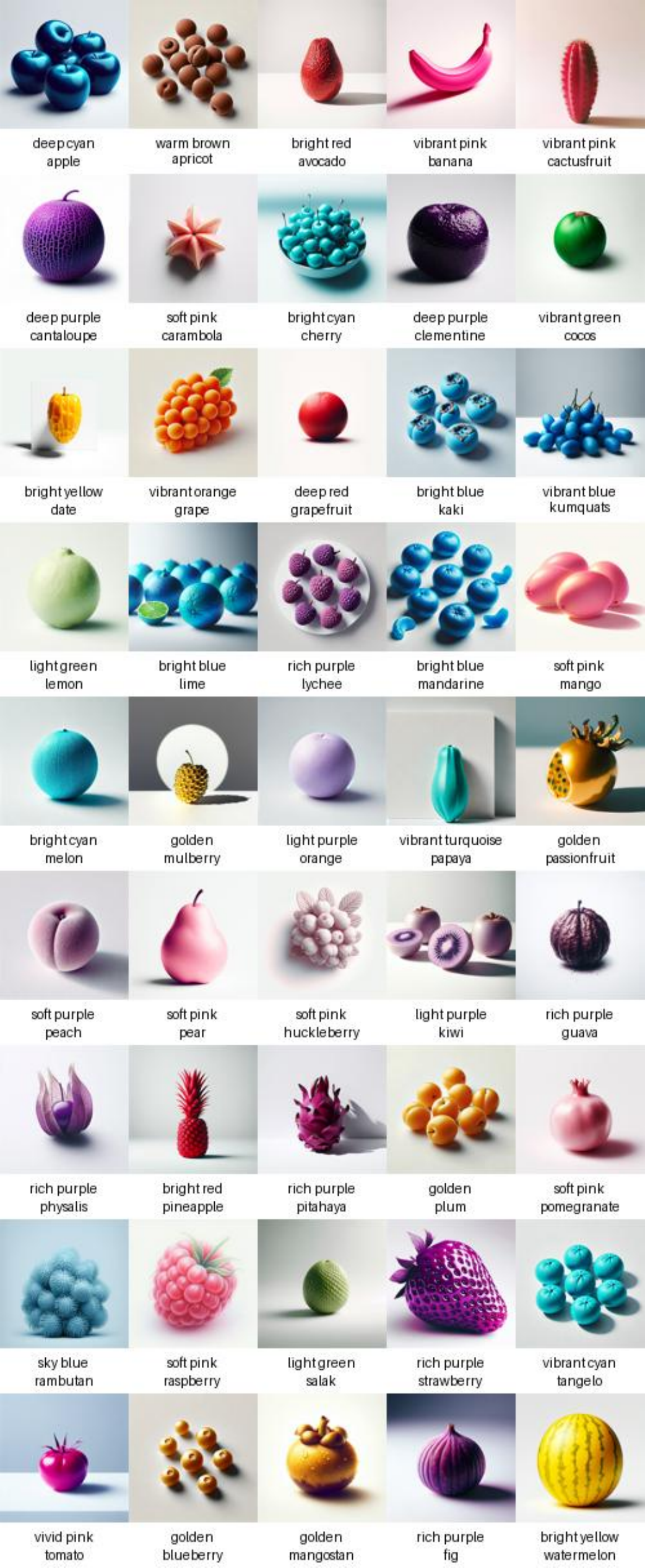}
\caption{Sample for each attribute-modified object in \method{}.}
\label{sup_fig_color}
\end{figure}

Examples of original objects and attribute-modified objects are shown in \cref{sup_fig_original,sup_fig_color}, respectively. All images in \method{} are standardized to a resolution of 1024$\times$1024.

\section{More Details on Model Evaluation}
\label{sup_sec_eval}

\subsection{Evaluation Settings}
For commercial models, we conduct experiments using CPU resources. 
For open-sourced models, we utilize a single RTX A6000 GPU (for models smaller than 26B) or multiple ones (for models larger than or equivalent to 26B) to evaluate models.

\subsection{Human Evaluation Details}

\begin{figure*}
\centering
\includegraphics[width=0.98\textwidth]{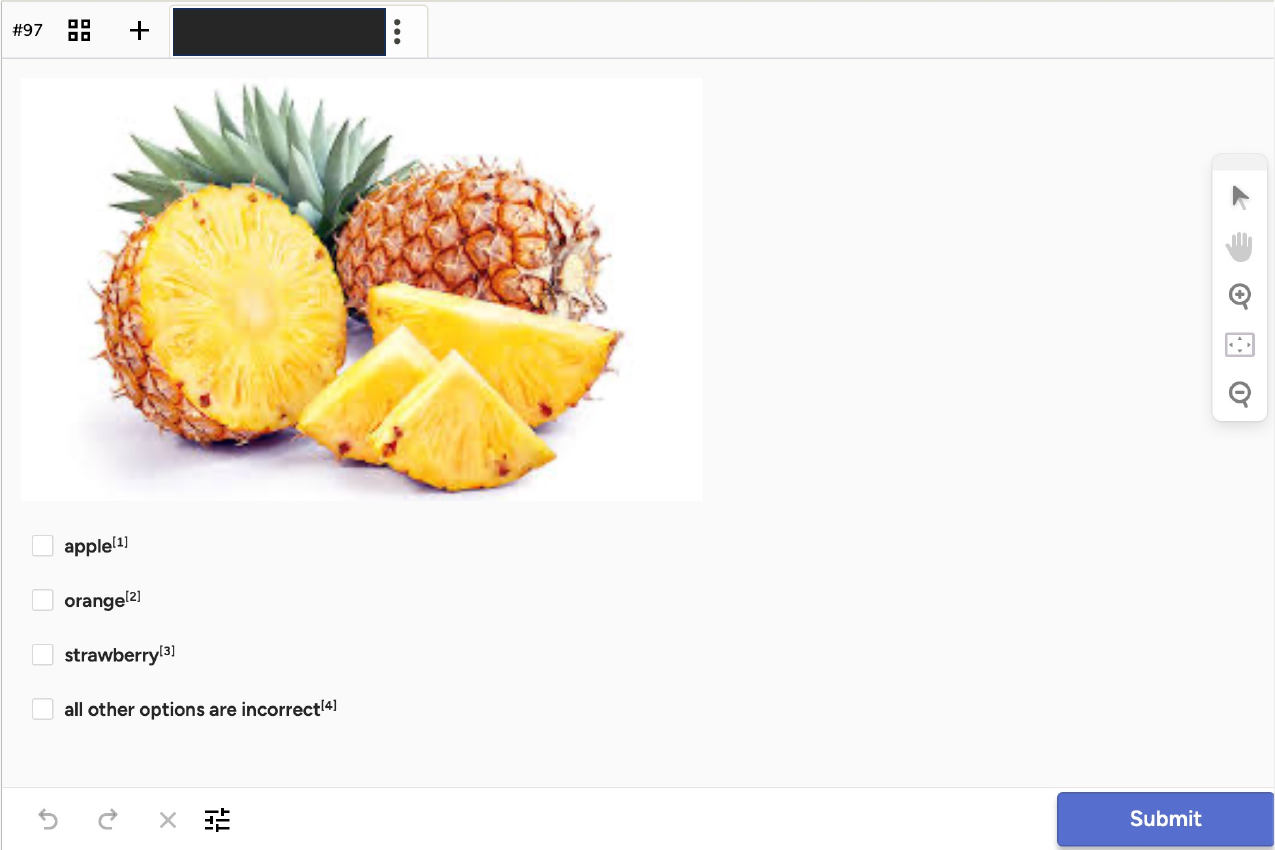}
\caption{Example of human evaluation interface.} 

\label{sup_fig_annotation}
\end{figure*}

\noindent
\textbf{Human annotators.}
We recruit two annotators to volunteer in our human evaluation process.

\noindent
\textbf{Human evaluation manual.}
We design three types of questions for evaluating original objects and attribute-modified objects. 
Ideally, each annotator would handle only one question type to ensure every image is evaluated once. However, due to limited numbers of annotators, both annotators evaluated all question types. To minimize potential evaluation bias, the order of image and question pairs is randomized, and these pairs are  not revisited by the annotators during the evaluation process.

\noindent
\textbf{Human evaluation interface.}
An example of the human evaluation interface is shown in \cref{sup_fig_annotation}. The interface is implemented using Label Studio~\cite{labelstudio}.

\subsection{Model Response Extraction}
\label{sup_subsec_extract}

As described in main paper Sec.~4,
for most models, we extract a single word or option directly from the model’s output. When necessary, we parse the output to obtain a comparable word or option.
Singular, plural forms, and synonyms of object names are also considered.

While for models including InstructBLIP~\cite{dai2023instructblip} (7B, 13B),  MiniGPT4~\cite{zhu2024minigpt} (7B, 13B), MiniGPT-v2-7B~\cite{chen2023minigptv2}, and Cambrian-1~\cite{tong2024cambrian} (3B, 8B, 13B, 34B),  which often produce sentences, we employ GPT-4o~\cite{gpt4v} assess alignment between the model’s response and the correct answer using the following prompt:

\begin{tcolorbox}[
  colframe=black,
  colback=gray!20,
  arc=2pt,
  left=0.2cm,
  right=0.2cm,
  top=2pt,
  bottom=2pt,
  boxrule=1pt,
  title=Prompt to Evaluate Model Responses via GPT-4o,
]

\small 
\texttt{Evaluate the following model response based on the question and correct answer:}

\noindent
\texttt{Question: \{question\}}

\noindent
\texttt{Correct Answer: \{correct\ answer\}}

\noindent
\texttt{Model Response: \{model\ response\}}

\noindent
\texttt{Is the model response correct? Please reply with "Correct" or "Incorrect".}
\end{tcolorbox}

A question is marked correct if the model's response matches the correct answer.

\subsection{Image-to-Image Matching}
In main paper Sec.~5.2 \textit{\textbf{Question 3}}, we employ image-to-image matching to investigate \textit{``Why does scaling up MLLM size not guarantee better performance?”} 
Here, we provide additional details on the image-to-image matching process.

To obtain the representation of each image, we compute the average embedding of all tokens within the image, as MLLMs omit the \texttt{[CLS]} token. 
The representations of the original objects form the index set. For each query image, we calculate the cosine similarity between the query and the indexed images to identify the Top-5 most similar images. 
In the Original-to-Original matching evaluation, we exclude the query image from the retrieved results to avoid trivial matches and ensure an unbiased evaluation.

\section{Model Failures}
\label{sup_sec_failure}

\subsection{Failure Cases for Open Questions}
\begin{figure}
\centering
\includegraphics[width=0.99\linewidth]{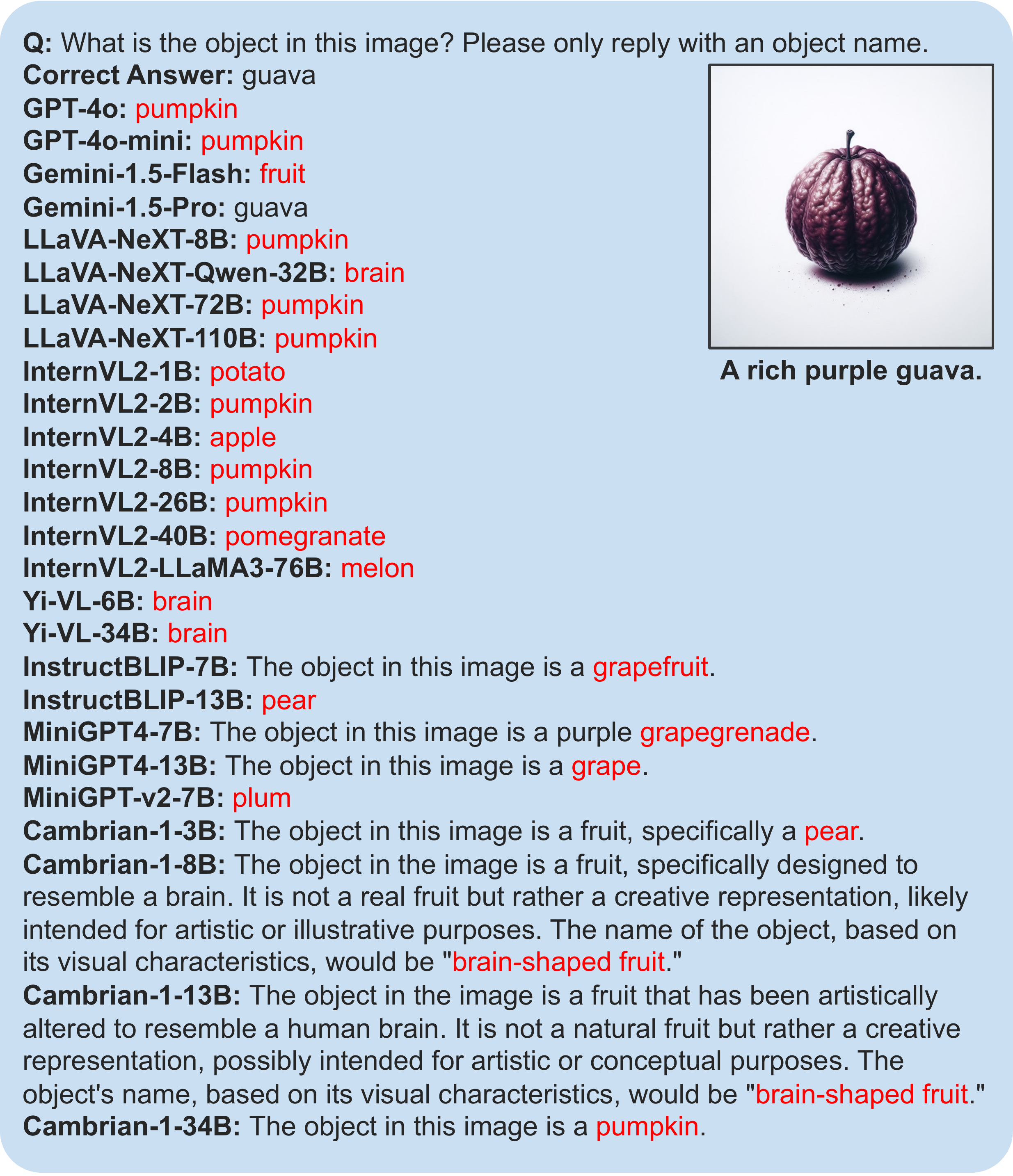}
\caption{Failures responses from 26 MLLMs for the open question on a attribute-modified sample of \textit{``A rich purple guava" } in \method{}.}
\label{sup_fig_failure1}
\end{figure}

\begin{figure}
\centering
\includegraphics[width=0.99\linewidth]{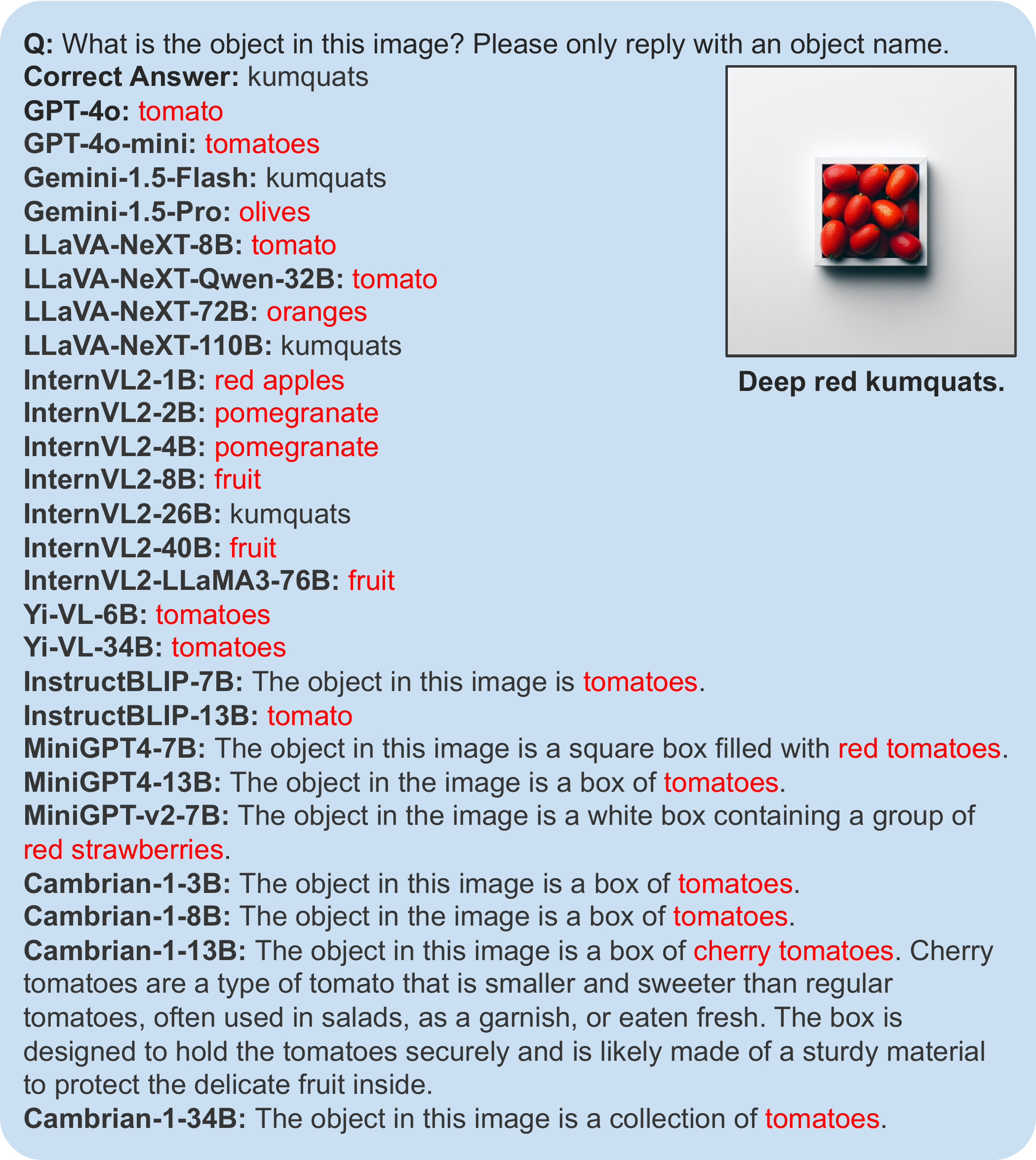}
\caption{Failures responses from 26 MLLMs for the open question on a attribute-modified sample of \textit{``deep red kumquats" } in \method{}.}
\label{sup_fig_failure2}
\end{figure}

\begin{figure}
\centering
\includegraphics[width=0.99\linewidth]{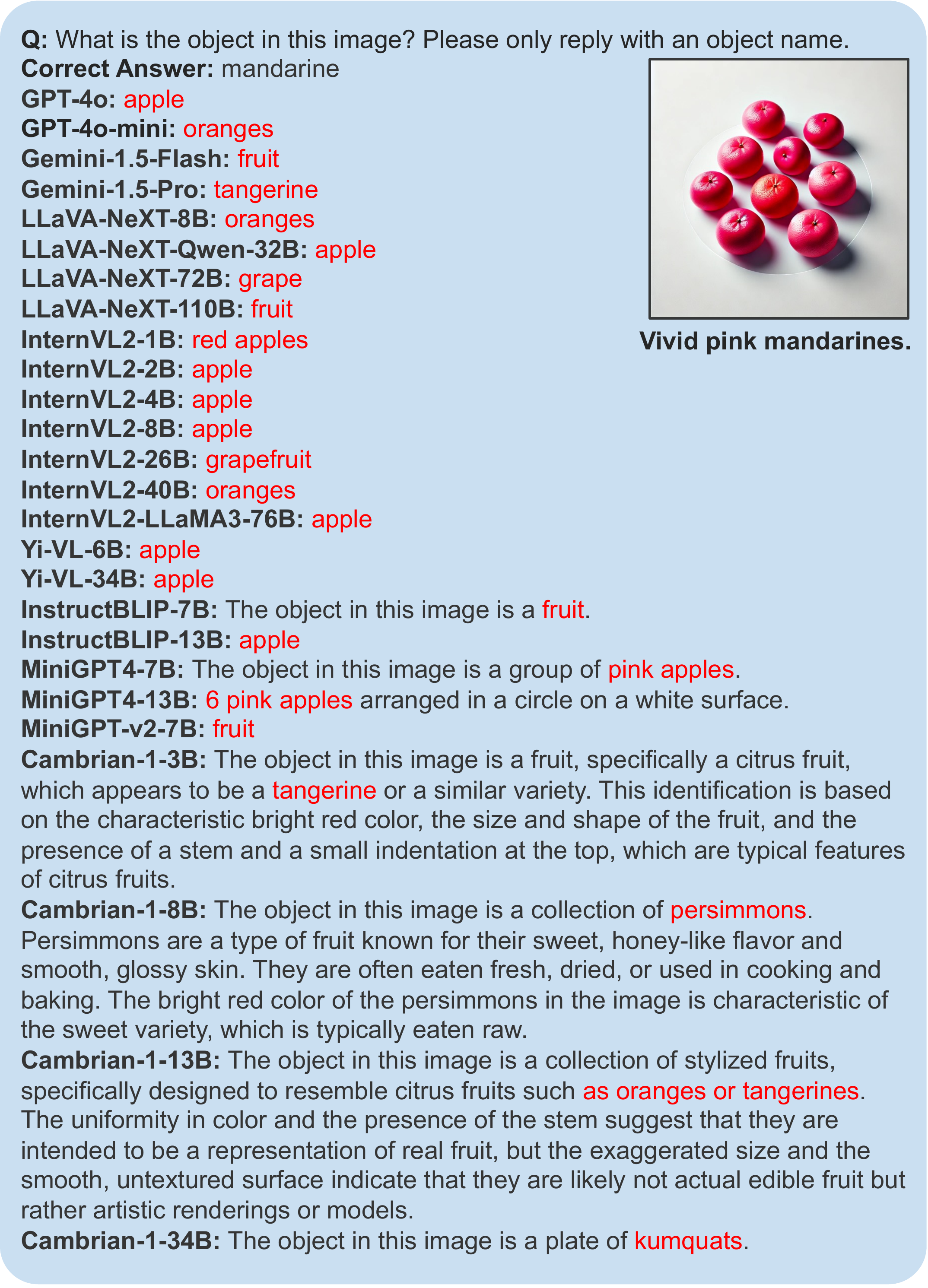}
\caption{Failures responses from 26 MLLMs for the open question on a attribute-modified sample of \textit{``vivid pink mandarines" } in \method{}.}
\label{sup_fig_failure3}
\end{figure}

We present examples of failure cases for MLLMs when answering open questions in \cref{sup_fig_failure1,sup_fig_failure2,sup_fig_failure3}, and identify several failure patterns as follows:

\noindent
\textbf{Hallucinations from spurious correlations.}
MLLMs often produce hallucinated responses due to spurious correlations between visual features. 
For instance, in \cref{sup_fig_failure1}, a \textit{``rich purple guava"} with a shape resembling a pumpkin is frequently misclassified as ``pumpkin" by most models. Similar issues occur in \cref{sup_fig_failure2} and \cref{sup_fig_failure3}: A \textit{"deep red kumquat"} is misidentified as a ``tomato" due to color similarity. A \textit{``vivid pink mandarine"} is mistaken for an ``apple" because of similar coloration, as well.
We further analyze such hallucinations in multiple-choice question settings in \cref{sup_subsec_hallu}.

\noindent
\textbf{Vague responses.}
Some MLLMs like Gemini-1.5-Flash in \cref{sup_fig_failure1}, InternVL2-8B, -40B, -LLaMA3-76B in \cref{sup_fig_failure2}, and LLaVA-NeXT-110B, InstructBLIP-7B, MiniGPT-v2-7B in  \cref{sup_fig_failure3}, fail to provide a specific object name when uncertain. 
Instead, they respond with general terms like ``fruit," which do not meet the requirement for precise object identification.
To address this, we design other question types including multiple-choice and unsolvable questions in the main paper, and also evaluate MLLMs with alternative open-question prompts in \cref{sup_subsec_other}.

\noindent
\textbf{Failure to interpret instructions.}
Some models fail to adhere to instruction prompts that specify concise responses. For example, models like InstructBLIP (7B, 13B), MiniGPT4 (7B, 13B), MiniGPT-v2-7B, and Cambrian-1 (3B, 8B, 13B, 34B) often generate full sentences instead of replying with a single object name. This behavior suggests these models struggle to interpret the instruction:
\textit{"What is the object in this image? Please only reply with an object name."}
To address this issue, we utilize GPT-4o to evaluate the alignment between these model outputs and correct answers, as described in \cref{sup_subsec_extract}.

\subsection{Hallucinations in Multiple-Choice Questions}
\label{sup_subsec_hallu}

\begin{figure*}
\centering
\includegraphics[width=0.99\textwidth]{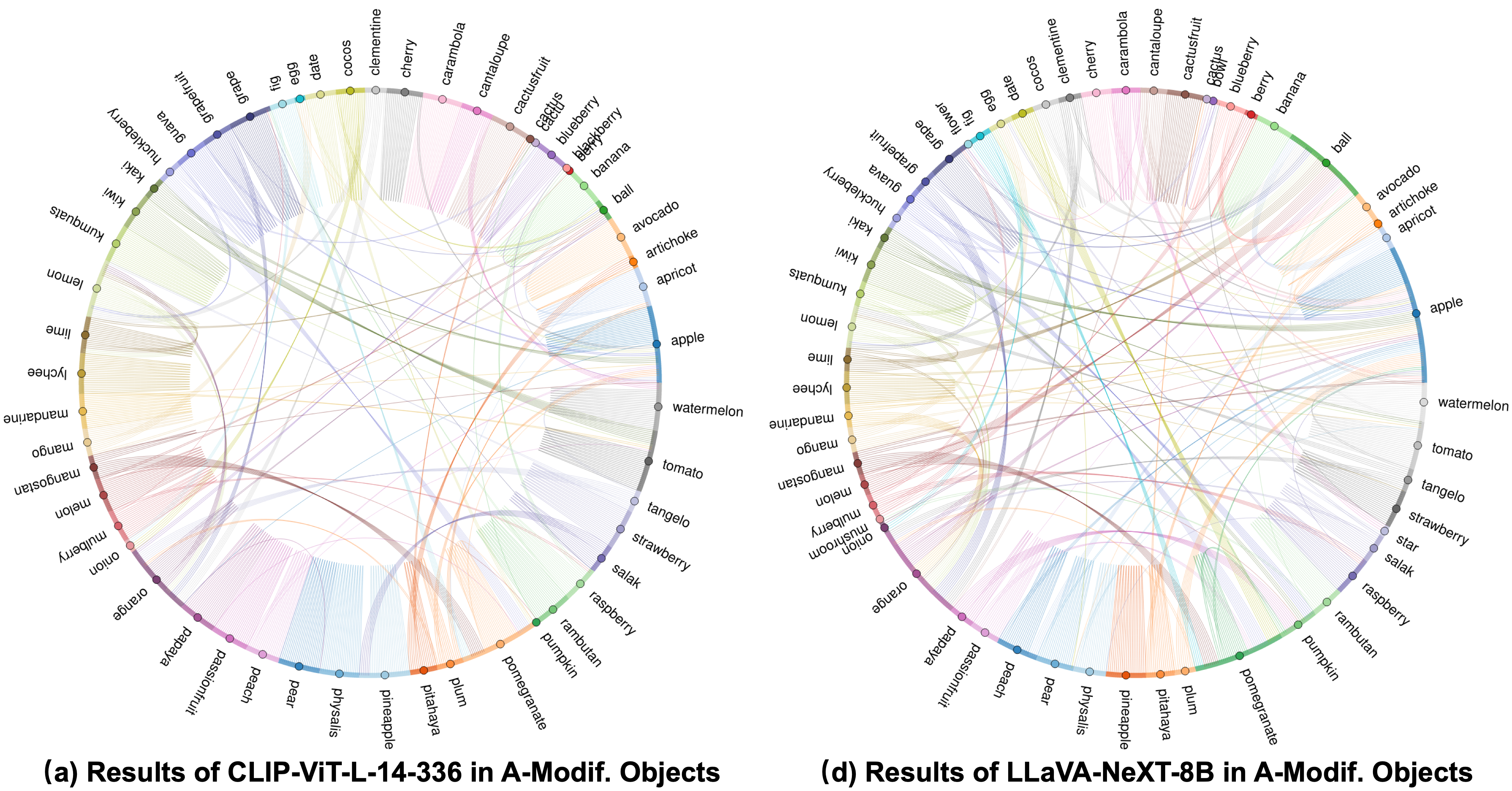}
\caption{Chord diagrams of LLaVA-NeXT-8B and its pre-trained vision encoder, CLIP-ViT-L-14-336, for multiple-choice questions on attribute-modified objects. The source node represents the correct object name. The target node represents the predicted object name. The edge color originates from the source node, indicating that the edge is linked from the correct object name to the predicted one.}
\label{sup_fig_chord}
\end{figure*}

In ~\cref{sup_fig_chord}, we visualize the results of LLaVA-NeXT-8B and its pre-trained vision encoder, CLIP-ViT-L-14-336, for multiple-choice questions on attribute-modified objects using chord diagrams. In the diagrams: the source node represents the correct object name. The target node represents the predicted object name. The edge color originates from the source node, indicating that the edge is linked from the correct object name to the predicted one. 
From the chord diagrams, we observe the following:

\noindent
\textbf{Increased Hallucinations in LLaVA-NeXT-8B.}
LLaVA-NeXT-8B exhibits significantly more hallucinations compared to its pre-trained vision encoder, CLIP-ViT-L-14-336. 
This observation further reinforces the conclusion from the main paper Sec.~5.2  \textit{\textbf{Question 1}} that the full potential of these encoders has yet to be fully realized in current MLLM models.

\noindent
\textbf{Frequent misclassification as common and easily recognizable objects like ``apple".}
A considerable number of objects are incorrectly classified as ``apple," suggesting a potential bias in the model towards common and easily recognizable objects.

These findings indicate that current MLLMs may introduce more hallucinations, especially for some  and easily recognizable objects, during the fine-tuning process.

\section{Further Analysis}
\label{sup_sec_analysis}

\subsection{Investigating on \method{}-Extension}

In main paper Sec.~5.4 \textit{\textbf{Question 5}} and \textit{\textbf{Question 6}}, we discuss the findings obtained from experiments on our \method{}-extension, which includes datasets of shape-modified fruits and color-modified animals. 
Below, we elaborate on experimental details and results.

\begin{figure}
\centering
\includegraphics[width=0.98\linewidth]{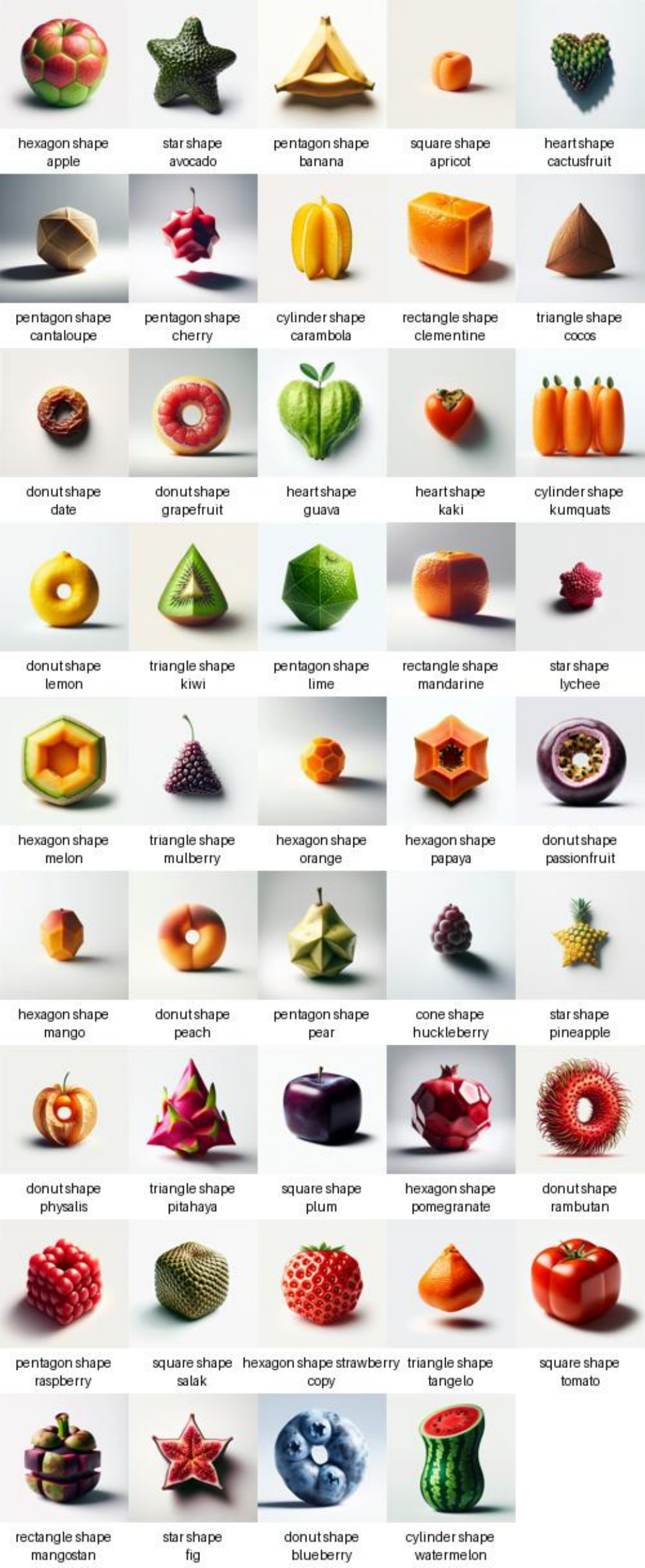}
\caption{Sample for each shape-modified object in \method{}-extension.}
\label{sup_fig_shape}
\end{figure}

\noindent
\textbf{Collection of shape-modified fruits.}
In addition to color modifications, we explored the performance of MLLMs on objects with other attribute changes by generating shape-modified images. Using DALL-E~\cite{Aditya2021dalle}, we generated images with prompts in the format, ``\textit{An image of \{shape\} \{object\}}", which resulted in a collection of 450 images with modified shapes, ultimately yielding 260 shape-modified images after selection. We show some examples in \cref{sup_fig_shape}, and the modified shapes for each object in \cref{suptab_fruit_shapes}.

\begin{table*}[htbp]
\centering
\caption{The modified shapes for each fruit in \method{}-extension.}
\label{suptab_fruit_shapes}
\resizebox{\linewidth}{!}{
\begin{NiceTabular}{@{}l|llllllllll@{}}
\toprule[1pt]
\textbf{Object} & \textbf{Shape$_1$} & \textbf{Shape$_2$} & \textbf{Shape$_3$} & \textbf{Shape$_4$} & \textbf{Shape$_5$} & \textbf{Shape$_6$} & \textbf{Shape$_7$} & \textbf{Shape$_8$} & \textbf{Shape$_9$} & \textbf{Shape$_{10}$}\\
\midrule
apple & triangle & hexagon & rectangle & pentagon & square & donut & cylinder & heart &  &   \\
apricot & square & triangle & hexagon & pentagon & heart &  &  &  &  &  \\
avocado & hexagon & star & cylinder &  &  &  &  &  &  &   \\
banana & square & triangle & donut & pentagon & heart & hexagon & star & cone &  & \\
blueberry & pentagon & square & rectangle & donut & star & cylinder &  &  &  &   \\
cactusfruit & pentagon & heart &  &  &  &  &  &  &  &  \\
cantaloupe & triangle & square & donut & star & pentagon & cone & cylinder &  &  & \\
carambola & square & hexagon & triangle & rectangle & pentagon & donut & star & cone & cylinder & heart \\
cherry & triangle & hexagon & square & heart & pentagon & donut & cylinder & star &  &   \\
clementine & triangle & square & pentagon & star & rectangle & donut & heart &  &  &  \\
cocos & triangle & rectangle &  &  &  &  &  &  &  & \\
date & square & donut & triangle & rectangle & cylinder &  &  &  &  &  \\
fig & square & triangle & heart & star & pentagon &  &  &  &  & \\
grapefruit & triangle & rectangle & hexagon & star & square & pentagon & triangle & heart & donut \\
guava & rectangle & square & hexagon & triangle & pentagon & heart &  &  &  &  \\
huckleberry & rectangle & square & donut & cone &  &  &  &  &  &   \\
kaki & square & heart & star &  &  &  &  &  &  &  \\
kiwi & square & triangle & rectangle & hexagon & donut & star & heart & cone & cylinder &  \\
kumquats & triangle & hexagon & cylinder &  &  &  &  &  &  &  \\
lemon & hexagon & triangle & pentagon & donut & star & heart &  &  &  &   \\
lime & donut & square & pentagon & star & heart & cone &  &  &  &  \\
lychee & triangle & rectangle & donut & star & hexagon & heart & pentagon &  &  & \\
mandarine & hexagon & rectangle & donut & pentagon & heart & star &  &  &  &  \\
mango & square & hexagon & pentagon & cone & cylinder & heart &  &  &  &   \\
mangostan & triangle & rectangle & donut &  &  &  &  &  &  &  \\
melon & square & triangle & hexagon & rectangle & donut & cone &  &  &  &    \\
mulberry & square & triangle & hexagon & donut & rectangle & heart & star & cylinder &  & \\
orange & square & hexagon & donut & heart & pentagon &  &  &  &  &  \\
papaya & rectangle & triangle & hexagon & square & pentagon & donut & star &  &  &  \\
passionfruit & triangle & donut & star & heart &  &  &  &  &  & \\
peach & hexagon & donut & star & rectangle & square & pentagon & heart &  &  &   \\
pear & pentagon & heart & donut &  &  &  &  &  &  &   \\
physalis & donut & star &  &  &  &  &  &  &  &  \\
pineapple & heart & star & donut &  &  &  &  &  &  &   \\
pitahaya & triangle & square & pentagon & star &  &  &  &  &  &  \\
plum & square & triangle & hexagon & donut & star & heart &  &  &  &   \\
pomegranate & square & triangle & rectangle & hexagon & pentagon & donut & star & heart &  &   \\
rambutan & square & rectangle & heart & donut &  &  &  &  &  &  \\
raspberry & square & triangle & pentagon & star & donut & heart & hexagon & cone & cylinder & \\
salak & square & hexagon & rectangle & pentagon & donut & triangle & heart & cone & star & cylinder   \\
strawberry & triangle & rectangle & square & donut & hexagon & pentagon & star & heart &  &  \\
tangelo & triangle & hexagon & square & rectangle & pentagon & star & donut & cylinder & heart &   \\
tomato & square & pentagon & star & rectangle &  &  &  &  &  & \\
watermelon & square & donut & rectangle & heart & star & cylinder &  &  &  &  \\

\bottomrule[1pt]
\end{NiceTabular}
}
\end{table*}

\noindent
\textbf{Collection of original animals and color-modified animals.}
To extend our analysis beyond fruits, we select 42 animal categories from ImageNet~\cite{deng2009imagenet}.
The generation process for attribute-modified images follows a similar approach in our main paper, using prompts in the format, ``\textit{An image of \{animal\} with \{color\} skin, replacing its usual skin}".
After filtering, we get 218 images of animals with modified colors. 
The modified colors for each animal are shown in \cref{suptab_animal_colors}.

\begin{figure}
\centering
\includegraphics[width=0.98\linewidth]{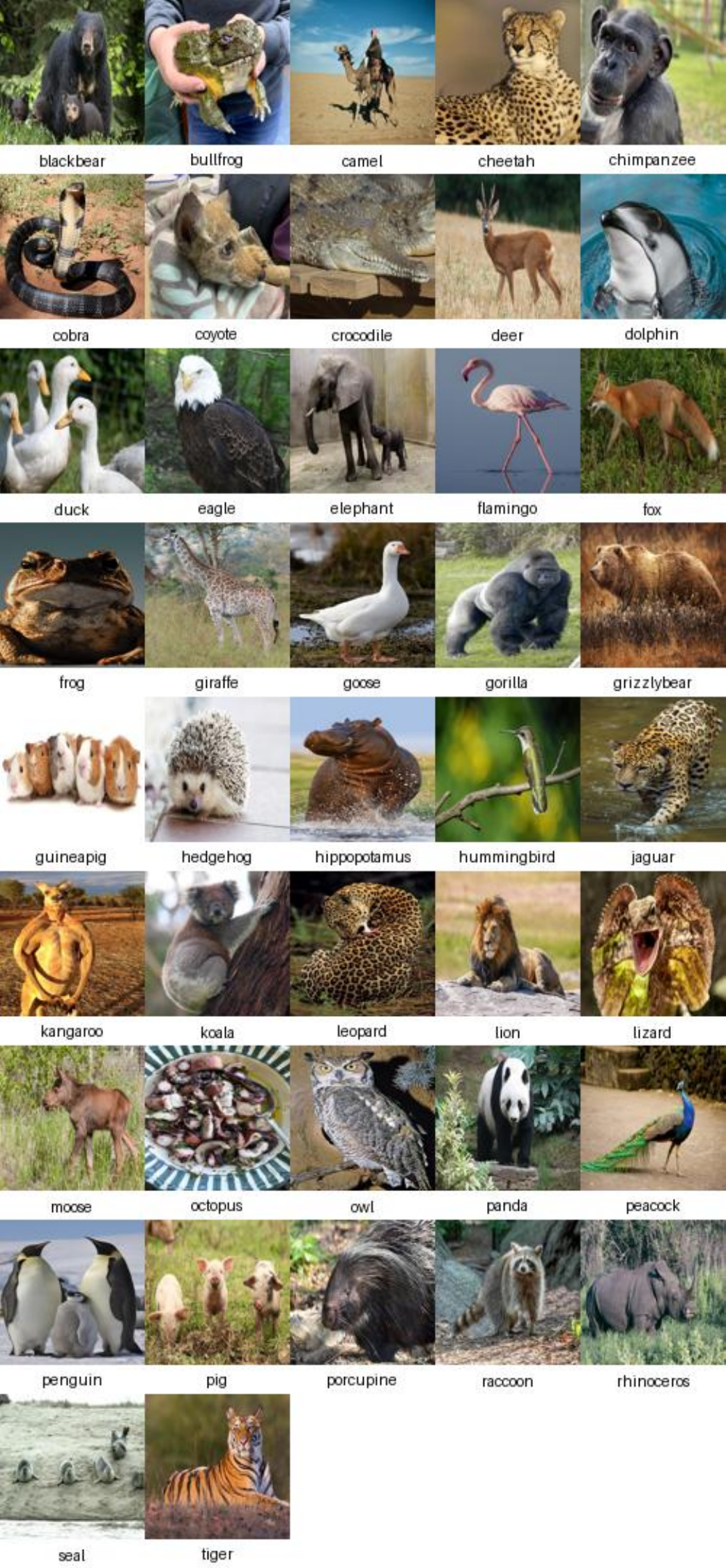}
\caption{Sample for each original animal in \method{}-extension.}
\label{sup_fig_animal_original}
\end{figure}

\begin{figure}
\centering
\includegraphics[width=0.98\linewidth]{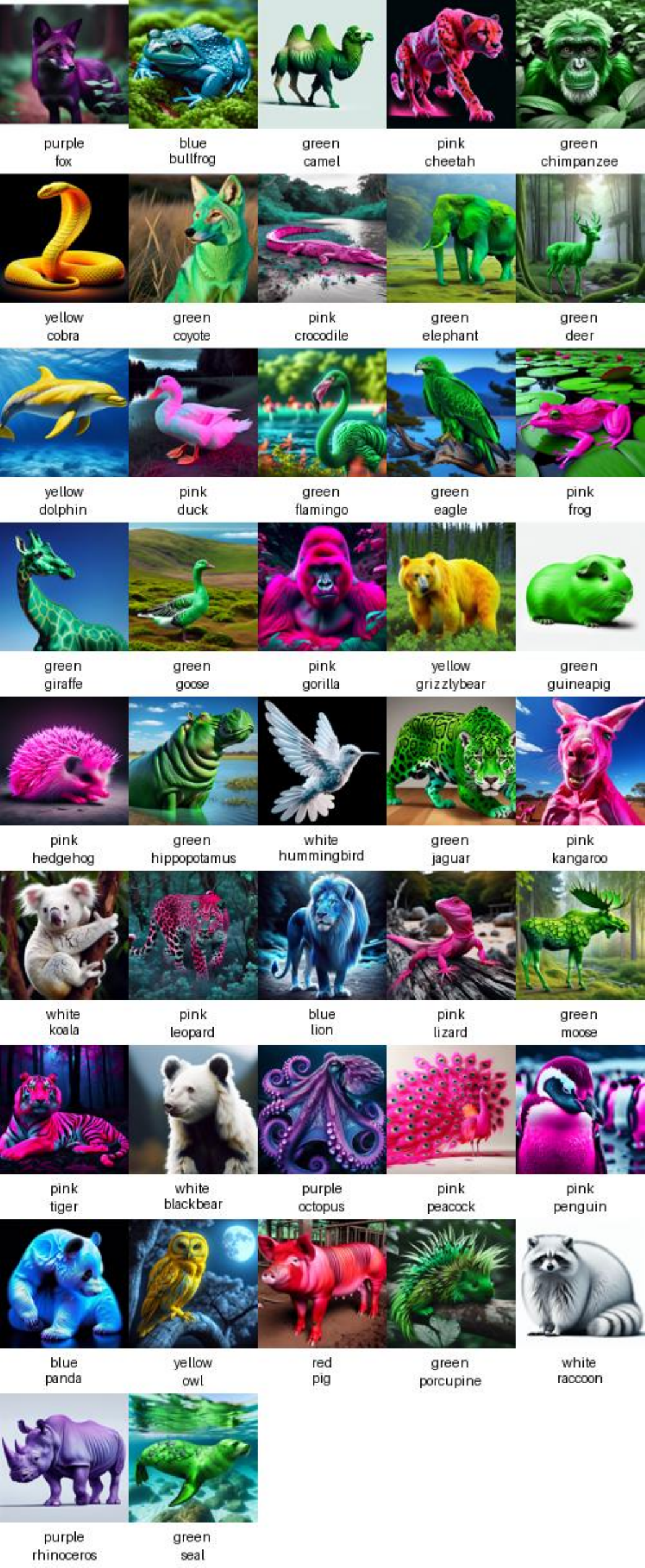}
\caption{Sample for each color-modified animal in \method{}-extension.}
\label{sup_fig_animal_color}
\end{figure}

To maintain consistency, we collected 218 original animal images from the Internet, matching the proportions of each animal category in the color-modified dataset. 
Examples of original and color-modified animals are presented in \cref{sup_fig_animal_original} and \cref{sup_fig_animal_color}, respectively.

\begin{table*}[htbp]
\centering
\caption{All animals and the modified colors for each animal in \method{}-extension.}
\label{suptab_animal_colors}
\setlength\tabcolsep{3pt}
\resizebox{\linewidth}{!}{
\begin{NiceTabular}{@{}l|llllllllll@{}}
\toprule[1pt]
\textbf{Animal} & \textbf{Color$_1$} & \textbf{Color$_2$} & \textbf{Color$_3$} & \textbf{Color$_4$} & \textbf{Color$_5$} & \textbf{Color$_6$} & \textbf{Color$_7$} & \textbf{Color$_8$} & \textbf{Color$_9$} & \textbf{Color$_{10}$} \\
\midrule
blackbear & neon pink & iridescent blue & pure white & deep purple & emerald green & sky blue & golden &  &  &  \\
bullfrog & bright yellow & neon pink & iridescent blue & pure white & fiery red & deep purple & sky blue & golden &  &  \\
camel & vibrant green & neon pink & iridescent blue & pure white & deep purple & emerald green & sky blue &  &  &  \\
cheetah & neon pink & iridescent blue & pure white & fiery red & deep purple & emerald green &  &  &  &  \\
chimpanzee & neon pink & pure white & fiery red & deep purple & emerald green &  &  &  &  &  \\
cobra & bright yellow & vibrant green & neon pink & iridescent blue & deep purple & emerald green & sky blue & golden &  &  \\
coyote & vibrant green & pure white & fiery red & deep purple & sky blue & golden &  &  &  &  \\
crocodile & vibrant green & neon pink & pure white & deep purple & emerald green &  &  &  &  &  \\
deer & bright yellow & vibrant green & neon pink & pure white &  &  &  &  &  &  \\
dolphin & bright yellow & vibrant green & pure white & emerald green &  &  &  &  &  &  \\
duck & bright yellow & neon pink &  &  &  &  &  &  &  &  \\
eagle & vibrant green & neon pink & deep purple & emerald green & golden &  &  &  &  &  \\
elephant & bright yellow & vibrant green & neon pink & iridescent blue & pure white & fiery red & deep purple & emerald green & sky blue & golden \\
flamingo & bright yellow & deep purple & emerald green & sky blue & golden &  &  &  &  &  \\
fox & bright yellow & neon pink & iridescent blue & deep purple & emerald green &  &  &  &  &  \\
frog & bright yellow & neon pink & pure white &  &  &  &  &  &  &  \\
giraffe & pure white & emerald green & golden &  &  &  &  &  &  &  \\
goose & bright yellow & vibrant green & neon pink &  &  &  &  &  &  &  \\
gorilla & bright yellow & neon pink &  &  &  &  &  &  &  &  \\
grizzlybear & bright yellow & vibrant green & neon pink & iridescent blue & fiery red &  &  &  &  &  \\
guineapig & bright yellow & vibrant green & neon pink & pure white  & deep purple & fiery red &  &  &  &  \\
hedgehog & vibrant green & neon pink & bright yellow & deep purple &  &  &  &  &  &  \\
hippopotamus & pure white & emerald green & sky blue &  &  &  &  &  &  &  \\
hummingbird & pure white & neon pink & deep purple & emerald green & golden & sky blue &  &  &  &  \\
jaguar & vibrant green & iridescent blue & neon pink & pure white & deep purple & emerald green & sky blue &  &  &  \\
kangaroo & neon pink & vibrant green & iridescent blue & fiery red & deep purple & emerald green & pure white &  &  &  \\
koala & neon pink & bright yellow & fiery red & pure white & deep purple &  &  &  &  &  \\
leopard & iridescent blue & fiery red & neon pink & sky blue & deep purple &  &  &  &  &  \\
lion & vibrant green & neon pink & iridescent blue & pure white & fiery red & sky blue &  &  &  &  \\
lizard & neon pink & pure white &  &  &  &  &  &  &  &  \\
moose & vibrant green & neon pink & iridescent blue & pure white & emerald green & sky blue &  &  &  &  \\
octopus & pure white & fiery red & deep purple & emerald green & sky blue & golden &  &  &  &  \\
owl & bright yellow & vibrant green & neon pink & iridescent blue & pure white & fiery red & deep purple & emerald green & sky blue &  \\
panda & neon pink & iridescent blue & deep purple &  &  &  &  &  &  &  \\
peacock & neon pink & deep purple & sky blue & golden &  &  &  &  &  &  \\
penguin & bright yellow & neon pink & pure white & fiery red & emerald green &  &  &  &  &  \\
pig & fiery red & deep purple & golden &  &  &  &  &  &  &  \\
porcupine & bright yellow & vibrant green & neon pink & iridescent blue & pure white &  &  &  &  &  \\
raccoon & pure white & fiery red &  &  &  &  &  &  &  &  \\
rhinoceros & vibrant green & neon pink & iridescent blue & deep purple & emerald green & golden &  &  &  &  \\
seal & bright yellow & vibrant green & neon pink & fiery red & golden &  &  &  &  &  \\
tiger & vibrant green & neon pink & fiery red & deep purple & emerald green &  &  &  &  &  \\

\bottomrule[1pt]
\end{NiceTabular}
}
\end{table*}

\noindent
\textbf{Results on \method{}-extension}
Tab.~\ref{sup_tab:other_overall} presents the results of MLLMs on objects with modified shapes. 
Note that, to compare fairly on the same data scale, the results for original fruits are reported as weighted averages derived from the main paper Sec.~5.1.  
These weights are calculated based on the proportion of each object category (e.g., \textit{``apple"}) relative to the total number of objects in the shape-modified fruit dataset.

The performance pattern observed here aligns closely with that seen for color-modified fruits in \method{}, demonstrating consistent trends across various types of attribute modifications and object categories.

\begin{table*}[tb]
\centering
\caption{Performance in terms of accuracy (\%) scores of 26 representative MLLMs on \method{}-extension, including open-sourced and commercial models, and random choice baseline. Higher score is better. 
The results cover original and shape-modified fruits, as well as original and color-modified animals.
\textbf{\colorbox{backblue1}{\makebox(18,5){\vspace{0.15mm}{Bold}}}} marks the best results, and \colorbox{backblue}{\makebox(24,5){\vspace{0.15mm}{original}}} marks the second best.}
\setlength\tabcolsep{3pt}
\label{sup_tab:other_overall}
\setlength\tabcolsep{3pt}
\resizebox{\linewidth}{!}{
\begin{NiceTabular}{@{}l|cccc:cccc|cccc:cccc@{}}
\CodeBefore
\rectanglecolor{softgray}{6-1}{9-17} 
\rectanglecolor{softgray}{17-1}{20-17} 
\rectanglecolor{softgray}{23-1}{24-17} 
\rectanglecolor{softgray}{29-1}{30-17} 
\Body
\toprule[1pt]
\multirow{2}{*}{\textbf{Method}} & \multicolumn{4}{c}{\textbf{Original Fruits}}  & \multicolumn{4}{c}{\textbf{Shape-Modified Fruits}}  & \multicolumn{4}{c}{\textbf{Original Animals}}  & \multicolumn{4}{c}{\textbf{Color-Modified Animals}} \\
\cmidrule(lr){2-5} \cmidrule(lr){6-9} \cmidrule(lr){10-13} \cmidrule(lr){14-17}
& \multicolumn{1}{c}{Open} & \multicolumn{1}{c}{Choice} & \multicolumn{1}{c}{Unsol.} & \multicolumn{1}{c}{Average} & \multicolumn{1}{c}{Open} & \multicolumn{1}{c}{Choice} & \multicolumn{1}{c}{Unsol.} & \multicolumn{1}{c}{Average}
& \multicolumn{1}{c}{Open} & \multicolumn{1}{c}{Choice} & \multicolumn{1}{c}{Unsol.} & \multicolumn{1}{c}{Average} & \multicolumn{1}{c}{Open} & \multicolumn{1}{c}{Choice} & \multicolumn{1}{c}{Unsol.} & \multicolumn{1}{c}{Average}\\
\midrule
\multicolumn{17}{l}{\textbf{Baselines}}\\
~~Random choice &-- &24.5\std2.2 &26.4\std2.0 &25.5 &--  &25.5\std2.4  &25.2\std2.1  &25.4  &-- &25.1\std3.0 &23.5\std2.1 &24.3 &-- &25.1\std3.3 &25.6\std2.5 &25.4 \\
\midrule

\multicolumn{17}{l}{\textbf{Open-Sourced Models}}  \\
~~LLaVA-NeXT-8B~\cite{li2024llavanext-strong} &42.5 &64.3\std0.6 &19.9\std13.5 &42.2 &28.5   &57.6\std0.6    &27.2\std16.4  &37.8 &57.8 &82.5\std1.4 &33.7\std11.6 &58.0 &58.3   &73.6\std1.3    &40.6\std18.7   &57.5  \\
~~LLaVA-NeXT-Qwen-32B~\cite{li2024llavanext-strong} &59.5 &83.4\std0.8 &39.2\std12.2 &60.7 &49.6   &81.8\std0.8    &\second{43.5}\std15.3   &58.3 &57.8 &93.2\std0.6 &37.5\std10.5 &62.8 &57.3   &83.0\std0.9    &\second{47.6}\std14.5   &62.6  \\
~~LLaVA-NeXT-72B~\cite{li2024llavanext-strong} &48.6 &73.1\std1.2 &26.7\std8.1 &49.5 &36.2   &72.0\std1.5    &31.9\std10.4   &46.7 &61.0  &89.0\std1.4  &26.6\std8.8 &66.8 &50.5   &79.5\std1.5   &36.5\std9.5   &55.5  \\
~~LLaVA-NeXT-110B~\cite{li2024llavanext-strong} &53.5 &76.5\std1.3 &8.3\std3.5 &46.1 &35.0  &75.8\std1.4    &14.5\std4.1   &41.8 &60.1 &87.2\std0.7 &12.5\std4.0 &53.3 &55.5   &76.5\std1.9    &13.3\std4.9   &48.4  \\

~~InternVL2-1B~\cite{chen2024far} &45.3 &66.2\std2.0 &13.1\std7.2 &41.5 &30.4   &67.6\std2.0    &21.4\std7.9   &39.8 &56.9 &81.0\std1.5 &18.7\std6.2 &52.2 &46.3   &75.4\std1.8    &12.2\std4.4   &44.6  \\
~~InternVL2-2B~\cite{chen2024far} &49.9 &67.1\std0.4 &14.2\std0.8 &43.7 &37.3   &56.6\std1.3    &8.6\std1.3   &34.2 &57.8 &80.3\std0.7 &14.6\std2.1 &50.9 &57.3   &74.1\std1.0    &7.7\std0.7   &46.4  \\
~~InternVL2-4B~\cite{chen2024far} &52.3 &71.0\std0.9 &15.2\std5.2 &46.2 &35.0  &71.0\std3.1 &15.8\std4.3   &40.6 &65.1  &88.5\std0.6 &8.2\std1.2  &53.9 &58.7   &76.8\std1.5    &6.4\std0.6   &47.3  \\
~~InternVL2-8B~\cite{chen2024far} &52.2 &85.0\std0.4 &31.6\std5.0 &56.3 &34.2 &67.7\std0.8  &30.0\std7.1  &44.0 &60.6 &88.8\std1.0 &31.0\std3.5 &60.1 &60.6   &81.1\std0.9    &29.1\std5.6 &57.3  \\
~~InternVL2-26B~\cite{chen2024far} &\best{77.7} &\best{95.4}\std0.7 &25.7\std8.3 &66.3 &\best{63.8}   &\second{86.2}\std0.7    &21.4\std5.8   &57.1 &71.1 &91.5\std1.1 &15.8\std2.5 &59.5 &68.8   &83.8\std1.6    &18.8\std3.1  &57.1  \\
~~InternVL2-40B~\cite{chen2024far} &68.1 &\second{94.7}\std0.5 &43.8\std8.8 &68.9 &51.2   &\best{86.3}\std1.5 &41.8\std6.5   &\second{59.8} &64.2 &91.8\std0.5 &32.4\std8.3 &62.8 &62.8   &84.3\std1.0   &34.0\std10.0   &60.4  \\
~~InternVL2-LLaMA3-76B~\cite{chen2024far} &\second{73.2} &75.6\std0.9 &31.2\std4.7 &60.0 &54.2   &73.4\std2.9 &31.2\std5.3   &52.9 &61.0 &87.6\std2.1 &19.5\std4.6 &56.0 &62.8   &79.9\std0.7  &17.8\std3.8   &53.5  \\

~~Cambrian-1-3B~\cite{tong2024cambrian} &61.9 &89.7\std0.9 &42.7\std7.7 &64.8 &46.9   &84.5\std0.6   &25.4\std3.9   &52.3 &90.4 &89.1\std1.0  &\second{44.6}\std3.3 &74.7 &\best{86.2}   &83.7\std0.8    &40.6\std2.2   &70.2  \\
~~Cambrian-1-8B~\cite{tong2024cambrian} &66.3 &87.3\std1.1 &\best{63.0}\std7.1 &\second{72.2} &57.7   &84.3\std0.8    &\best{45.3}\std3.9   &\best{62.4} &\second{93.6} &\best{94.4}\std0.8  &\best{49.3}\std4.5  &\best{79.1} &\second{82.6}  &\best{86.7}\std1.1   &\best{48.3}\std1.5     &\best{72.5}  \\
~~Cambrian-1-13B~\cite{tong2024cambrian} &62.8 &85.7\std0.3 &44.6\std6.5 &64.4 &\second{61.5}   &81.6\std0.8    &34.1\std2.6   &59.1 &92.7 &\second{93.6}\std1.8 &40.1\std4.0 &75.5 &77.1  &\second{86.5}\std2.8    &39.8\std2.5   &67.8  \\
~~Cambrian-1-34B~\cite{tong2024cambrian} &69.5 &91.3\std1.3 &\second{57.0}\std2.3 &\best{72.6} &56.2   &86.0\std1.2    &33.8\std3.8    &58.7 &\best{94.5}  &91.6\std1.4 &44.4\std4.3 &\second{76.8} &\best{86.2}  &84.5\std1.1 &40.3\std3.3 &\second{70.3}  \\

~~Yi-VL-6B~\cite{young2024yi} &54.1 &87.8\std1.0 &42.5\std8.0 &61.5 &40.4   &79.9\std2.1 &37.9\std10.0 &52.7 &61.0 &81.3\std1.2 &37.4\std8.6 &59.9 &59.2 &73.5\std0.8 &34.3\std11.0 &55.7 \\
~~Yi-VL-34B~\cite{young2024yi} &54.0 &90.7\std1.2 &26.5\std9.4 &57.1 &36.9  &74.4\std1.2   &27.1\std7.7    &46.1 &67.0  &83.5\std1.0 &21.3\std12.5 &57.3 &57.3  &79.6\std1.5 &37.1\std12.4 &58.0 \\

~~InstructBLIP-7B~\cite{dai2023instructblip}  &58.7 &43.0\std10.9 &5.8\std4.3 &35.8 &46.5   &48.5\std10.1    &14.4\std18.0  &36.5 &69.7 &58.6\std56.3 &12.8\std10.6  &47.0 &63.8   &54.0\std8.7    &13.9\std8.7   &43.9 \\
~~InstructBLIP-13B~\cite{dai2023instructblip} &57.8 &56.0\std11.6 &5.8\std6.0 &39.9 &38.8   &57.0\std11.2    &37.8\std33.3   &44.5 &65.6 &69.2\std11.1 &35.1\std37.1 &56.6 &61.0   &64.1\std8.6    &39.2\std35.3  &54.8 \\
~~MiniGPT4-7B~\cite{zhu2024minigpt} &56.4 &59.1\std1.7 &35.9\std3.7 &50.5 &44.2   &30.0\std5.4    &7.2\std3.5  &27.1 &83.9 &68.1\std1.1 &6.4\std1.2 &52.8 &75.2   &48.3\std2.1    &7.3\std2.3 &43.6 \\
~~MiniGPT4-13B~\cite{zhu2024minigpt}  &56.4 &58.9\std2.9 &39.9\std0.9 &51.7 &44.6   &45.3\std0.9    &7.1\std3.5   &32.3 &89.4 &76.6\std1.3 &5.0\std2.8 &57.0 &72.9   &65.9\std1.1    &6.0\std2.6   &48.3 \\
~~MiniGPT4-v2-7B~\cite{chen2023minigptv2} &55.5 &58.7\std4.5 &30.1\std20.4 &48.1 &35.4   &49.9\std4.9   &17.4\std25.6  &34.2 &69.7 &74.6\std3.5 &6.0\std8.8 &50.1 &61.5   &68.8\std3.6    &9.0\std15.2   &46.4 \\

\midrule

\multicolumn{17}{l}{\textbf{Commercial Models}} \\
~~GPT-4o~\cite{gpt4v}  &\best{88.1} &\second{93.3}\std0.4 &\best{78.8}\std3.4 &\best{86.7} &61.5   &88.7\std0.3   &86.2\std6.4   &\best{78.8} &78.4 &\best{97.9}\std0.5 &\second{82.4}\std2.7 &\second{86.2} &\second{71.1}  &\second{90.8}\std0.7     &\second{82.1}\std7.0   &\best{81.3} \\
~~GPT-4o-mini~\cite{gpt4v} &79.1 &88.0\std0.1 &\second{78.0}\std1.9 &\second{81.7} &61.5    &85.6\std1.1   &\second{82.2}\std3.1   &76.4  &79.4 &\second{97.4}\std0.5 &\best{82.5}\std1.7 &\best{86.4} &70.6    &87.9\std0.8  &\best{84.5}\std3.7   &\second{81.0} \\
~~Gemini-1.5-Flash~\cite{gemini} &\second{86.6} &92.9\std0.1 &63.7\std3.4 &81.1 &\second{71.2}    &\second{89.6}\std1.1   &43.8\std15.5   &68.2 &75.7 &95.0\std1.2 &22.3\std7.3 &64.3 &70.2   &89.9\std1.1   &24.8\std8.1   &61.6 \\
~~Gemini-1.5-Pro~\cite{gemini} &84.9 &\best{94.7}\std0.6 &56.0\std10.7 &78.5 &\best{77.3}   &\best{94.1}\std0.9 &58.8\std8.0   &\second{76.7}  &\best{79.8} &96.8\std0.6 &30.3\std14.6 &68.9 &\best{72.9}   &\best{92.4}\std2.2   &26.5\std10.9  &63.9 \\
\bottomrule[1pt]
\end{NiceTabular}
}
\end{table*}

\subsection{Investigating on Synthetic Original Data}
To ensure a fair comparison, we also synthesize original objects using DALL-E~\cite{Aditya2021dalle}. Specifically, we generated images with the prompt \textit{``An image of \{object\}.”} to evaluate whether the use of generated images introduces performance degradation due to the generation model.

As shown in \cref{sup_tab_syn}, MLLMs' performance on synthetic original data exhibits a slight drop of approximately 5 points in average accuracy, which is considerably smaller than the drop observed for attribute-modified objects. These results suggest that the use of generated images has minimal negative impact on the performance of MLLMs, confirming that synthetic data from the generation model is a relatively reliable approach for evaluating their performance on attribute-modified objects.

\begin{table*}[tb]
\centering
\caption{Performance in terms of accuracy (\%) scores of representative MLLMs on original objects from \method{} and synthetic original objects. Higher score is better.
\textbf{\colorbox{backblue1}{\makebox(18,5){\vspace{0.15mm}{Bold}}}} marks the best results, and \colorbox{backblue}{\makebox(24,5){\vspace{0.15mm}{original}}} marks the second best.}
\label{sup_tab_syn}
\setlength\tabcolsep{3pt}
\resizebox{0.75\linewidth}{!}{
\begin{NiceTabular}{@{}l|cccc|cccc@{}}
\CodeBefore
\rectanglecolor{softgray}{4-1}{7-9} 
\rectanglecolor{softgray}{16-1}{17-9} 
\Body
\toprule[1pt]
\multirow{2}{*}{\textbf{Method}} 
& \multicolumn{4}{c|}{\textbf{Original Objects}} 
& \multicolumn{4}{c}{\textbf{Synthetic Original Objects}} \\
\cmidrule(lr){2-5} \cmidrule(lr){6-9}
& \multicolumn{1}{c}{Open} 
& \multicolumn{1}{c}{Choice} 
& \multicolumn{1}{c}{Unsol.} 
& \multicolumn{1}{c|}{Average}
& \multicolumn{1}{c}{Open} 
& \multicolumn{1}{c}{Choice} 
& \multicolumn{1}{c}{Unsol.} 
& \multicolumn{1}{c}{Average}\\
\midrule

\multicolumn{8}{l}{\textbf{Open-Sourced Models}}\\
~~LLaVA-NeXT-8B~\cite{li2024llavanext-strong} & 42.0 & 77.2\std0.8 & 27.1\std13.9 & 48.8 &41.6 &64.9\std0.7 &21.0\std14.0 &42.5  \\
~~LLaVA-NeXT-Qwen-32B~\cite{li2024llavanext-strong} & 58.0 & 93.8\std0.3 & \second{42.0}\std16.5 & 64.6 &56.0 &84.6\std0.7 &\second{39.9}\std12.8 &60.2 \\
~~LLaVA-NeXT-72B~\cite{li2024llavanext-strong} & 47.6 & 84.5\std0.5 & 28.2\std9.7 & 53.4  &45.1 &73.6\std1.3 &30.0\std8.3 &49.6  \\
~~LLaVA-NeXT-110B~\cite{li2024llavanext-strong} & 51.3 & 85.5\std1.2 & 16.2\std8.3 & 51.0  &47.8 &77.7\std1.0 &9.6\std5.0 &45.0  \\

~~InternVL2-1B~\cite{chen2024far} & 45.3 & 71.5\std1.0 & 21.6\std10.4 & 46.1  &42.2  &64.7\std1.6 &14.3\std7.7 &40.4  \\
~~InternVL2-2B~\cite{chen2024far} & 49.6 & 68.4\std0.5 & 13.8\std0.4 & 43.9  &46.4 &65.8\std1.5 &25.0\std25.4 &45.7  \\
~~InternVL2-4B~\cite{chen2024far} & 50.9 & 82.9\std0.7 & 18.6\std6.0 & 50.8  &42.9 &70.9\std0.9 &16.8\std5.0 &43.5  \\
~~InternVL2-8B~\cite{chen2024far} & 50.7 & 84.3\std0.5 & 31.3\std5.2 & 55.4  &44.2 &73.7\std0.8 &28.1\std6.9 &48.7  \\
~~InternVL2-26B~\cite{chen2024far} & \best{79.3} & \best{96.0}\std0.5 & 24.3\std8.1 & \second{66.5} &\best{74.4}  &\second{87.9}\std1.6 &23.8\std5.6 &\second{62.0} \\
~~InternVL2-40B~\cite{chen2024far} & 69.6 & \second{95.4}\std0.3 & \best{44.9}\std8.6 & \best{70.0}  &\second{65.1} &\best{92.3}\std1.4 &\best{53.3}\std8.6 &\best{70.2}  \\
~~InternVL2-LLaMA3-76B~\cite{chen2024far} & \second{73.3} & 87.0\std0.9 & 35.7\std6.1 & 65.3  &64.7 &78.0\std0.9 &33.4\std5.4 &58.8  \\

\midrule
\multicolumn{8}{l}{\textbf{Commercial Models}} \\
~~GPT-4o~\cite{gpt4v} & \best{90.9} & \best{99.3}\std0.2 & \second{84.2}\std1.9 & \best{91.5}  &\second{86.2} &\second{93.8}\std0.3 &\second{79.4}\std2.9 &\best{86.5}  \\
~~GPT-4o-mini~\cite{gpt4v} & 85.3 & 95.4\std0.4 & \best{87.0}\std1.1 & \second{89.2}  &81.3 &88.8\std0.1 &\best{79.9}\std1.7 &\second{83.3}  \\
~~Gemini-1.5-Flash~\cite{gemini} & 86.2 & 96.1\std1.2 & 57.9\std15.0 & 80.1  &\best{87.6} &93.7\std0.1 &66.2\std3.2 &82.5  \\
~~Gemini-1.5-Pro~\cite{gemini} & \second{86.4} & \second{99.2}\std0.4 & 68.3\std1.0 & 84.6  &85.1 &\best{94.9}\std0.5 &58.0\std10.7 &79.3  \\

\bottomrule[1pt]
\end{NiceTabular}
}
\end{table*}

\subsection{Investigating on Other Question Prompts}
\label{sup_subsec_other}

\noindent
\textbf{Open questions using other prompts.}
In the main paper, we primarily used the prompt \textit{``What is the object in this image? Please only reply with an object name.”} to evaluate the model's performance in open questions. In this section, we experiment with alternative prompts, such as \textit{``What is the fruit in this image? Please only reply with a fruit name.”}, which specifies the object category as fruit.

\noindent
\textbf{Open questions without limitation.}
We further examine the model's performance with prompts like \textit{``What is the object in this image?"} and \textit{``What is the fruit in this image?"}, allowing the model to provide responses of any length without restricting it to a single word.

As summarized in \cref{sup_tab_open}, the experimental results reveal that variations in open question phrasing lead to marginal differences in accuracy. However, a noticeable performance drop persists on attribute-modified objects. 
These findings highlight the relative robustness of MLLMs to prompt phrasing variations within this evaluation context, while also emphasizing the challenges posed by attribute-modified objects.

\begin{table*}[tb]
\centering
\caption{Performance in terms of accuracy (\%) scores of representative MLLMs on \method{}. Higher score is better. 
The evaluation spans original and synthetic original objects across four open question types: open questions requiring an object name as the response (Obj. Nam.); open questions requiring a fruit name as the response (Fru. Nam.); open questions with unrestricted object responses of any length (Obj. Unre.), and (4) open questions with unrestricted fruit responses of any length (Fru. Unre.).
\textbf{\colorbox{backblue1}{\makebox(18,5){\vspace{0.15mm}{Bold}}}} marks the best results, and \colorbox{backblue}{\makebox(24,5){\vspace{0.15mm}{original}}} marks the second best.}
\label{sup_tab_open}
\setlength\tabcolsep{3pt}
\resizebox{0.78\linewidth}{!}{
\begin{NiceTabular}{@{}l|cccc|cccc@{}}
\CodeBefore
\rectanglecolor{softgray}{4-1}{7-9} 
\rectanglecolor{softgray}{15-1}{16-9} 
\Body
\toprule[1pt]
\multirow{2}{*}{\textbf{Method}} 
& \multicolumn{4}{c|}{\textbf{Original Objects}} 
& \multicolumn{4}{c}{\textbf{Attribute-Modified Objects}} \\
\cmidrule(lr){2-5} \cmidrule(lr){6-9}
& \multicolumn{1}{c}{Obj. Nam.} 
& \multicolumn{1}{c}{Fru. Nam.} 
& \multicolumn{1}{c}{Obj. Unre.} 
& \multicolumn{1}{c|}{Fru. Unre.}
& \multicolumn{1}{c}{Obj. Nam.} 
& \multicolumn{1}{c}{Fru. Nam.} 
& \multicolumn{1}{c}{Obj. Unre.} 
& \multicolumn{1}{c}{Fru. Unre.}\\
\midrule

\multicolumn{8}{l}{\textbf{Open-Sourced Models}}\\
~~LLaVA-NeXT-8B~\cite{li2024llavanext-strong} &42.0 &54.0 &58.7 &59.3  &24.9 &34.2 &26.7 &32.2  \\
~~LLaVA-NeXT-Qwen-32B~\cite{li2024llavanext-strong}&58.0 &\second{66.9} &61.1  &75.8  &35.8 &47.3 &43.1 &\second{52.9}  \\
~~LLaVA-NeXT-72B~\cite{li2024llavanext-strong} &47.6 &50.9 &48.9 &59.8  &24.2 &26.9 &23.3 &26.7  \\
~~LLaVA-NeXT-110B~\cite{li2024llavanext-strong} &51.3 &52.9 &55.6 &53.6  &26.9 &30.0 &24.0 &28.4  \\

~~InternVL2-1B~\cite{chen2024far} &45.3 &43.1 &38.7 &40.4  &24.0 &29.3 &19.8 &22.4  \\
~~InternVL2-2B~\cite{chen2024far} &49.6 &51.3 &52.9 &54.7  &25.6 &30.0 &22.9 &27.3  \\
~~InternVL2-4B~\cite{chen2024far} &50.9 &48.7 &44.2 &50.9  &23.3 &28.9 &19.6 &23.8  \\
~~InternVL2-8B~\cite{chen2024far} &50.7 &56.7 &57.6 &61.6  &25.8 &28.7 &23.8 &31.3  \\ 
~~InternVL2-26B~\cite{chen2024far} &\best{79.3} &\best{75.6} &\second{75.3} &\best{81.6}  &\best{53.3} &\best{56.9} &\best{49.1} &\best{54.0}  \\
~~InternVL2-LLaMA3-76B~\cite{chen2024far} &\second{73.3} &\best{75.6} &\best{79.8} &\second{80.2}  &\second{44.4} &\second{50.0} &\second{42.2} &46.9  \\

\midrule
\multicolumn{8}{l}{\textbf{Commercial Models}} \\
~~GPT-4o~\cite{gpt4v} &\best{90.9} &\best{92.4} &87.8 &\second{92.0}  &56.2 &62.0 &47.8 &58.7  \\
~~GPT-4o-mini~\cite{gpt4v} &85.3 &88.4 &86.0 &91.8  &49.6 &59.1 &43.8 &54.4  \\
~~Gemini-1.5-Flash~\cite{gemini} &86.2 &\second{92.0} &\second{88.0} &91.1  &\best{63.3} &\best{69.1} &\best{56.2} &\second{63.1}  \\
~~Gemini-1.5-Pro~\cite{gemini} &\second{86.4} &88.9 &\best{89.1} &\best{92.7}  &\second{62.2} &\second{67.3} &\second{56.0} &\best{63.3}  \\

\bottomrule[1pt]
\end{NiceTabular}
}
\end{table*}

\section{More Discussion}
\label{sup_sec_discuss}

\subsection{Uniqueness of \method}
\label{sup_subsec:uniqueness}

\method{} is designed to evaluate MLLMs' capabilities in object recognition, reasoning from commonsense to beyond it, by introducing attribute-modified objects alongside original ones.
This new benchmark is unique for two main reasons:

\noindent
\textbf{Reasoning from commonsense to beyond commonsense.}
\method{} bridges commonsense recognition with beyond-commonsense reasoning by incorporating both original and attribute-modified objects.
It evaluates the model’s transition from recognizing original objects to understanding uncommon, modified variations, reflecting a more comprehensive and nuanced measure of MLLMs' flexibility.

\noindent
\textbf{Multi-question designing.} 
\method{} employs a well-constructed question design with three distinct question types: open questions, multiple-choice questions, and unsolvable questions.
This setup assesses the model’s ability to identify objects, differentiate between correct and distractor options, and recognize when no correct answer is available, offering a comprehensive evaluation of the model's strengths and limitations.

\subsection{Is \method{} easy?}

Yes, \method{} is easy for humans. As demonstrated by the results in the main paper Sec.~5.1, humans can readily identify attribute-modified objects using their prior knowledge. 
However, through our experiments, we find \method{} poses significant challenges for current MLLMs due to the following reasons:

\noindent
\textbf{From the model perspective.}
On the one hand, the full generalization capabilities of vision encoders are not fully realized in current MLLMs.
On the other hand, larger LLMs often degrade the effectiveness of vision encoders during fine-tuning, leading to suboptimal integration.

\noindent
\textbf{From the data perspective.} 
Spurious correlations between visual features hinder MLLMs' ability to accurately understand attribute-modified objects in \method{}.

Therefore, our \method{} can serve as as a testbed to lay out potential roadmaps to enhance versatile and resilient MLLMs.

\end{document}